\documentclass[lettersize,journal]{IEEEtran}
\usepackage{amsmath,amsfonts}
\usepackage{algorithmic}
\usepackage{algorithm}
\usepackage{array}
\usepackage[caption=false,font=normalsize,labelfont=sf,textfont=sf]{subfig}
\usepackage{textcomp}
\usepackage{stfloats}
\usepackage{url}
\usepackage{verbatim}
\usepackage{graphicx}
\usepackage{cite}
\usepackage[T1]{fontenc}
\usepackage{times}
\usepackage{helvet}
\usepackage{amssymb}
\usepackage{multirow}
\usepackage{cite}

\usepackage{pgfplots} 
\pgfplotsset{compat=newest}
\usepackage{color}
\definecolor{forestgreen}{RGB}{0,139,69}

\usepackage{xcolor}
\definecolor{citecolor}{HTML}{0071bc}
\definecolor{SeaGreen4}{RGB}{0,205,102} 
\definecolor{SlateBlue}{RGB}{106,90,205} 
\definecolor{DarkRed}{RGB}{178,34,34} 

\usepackage[switch]{lineno}

\usepackage{textcomp,booktabs}
\usepackage{amssymb}
\usepackage{pifont}

\usepackage{makecell}

\usepackage{colortbl}
\definecolor{mygray}{gray}{.9}
\definecolor{mypink}{rgb}{.99,.91,.95}
\definecolor{mycyan}{cmyk}{.3,0,0,0}

\usepackage[colorlinks, linkcolor=red,  anchorcolor=blue, citecolor=citecolor]{hyperref}

\begin{document}

\title{An Empirical Study of Mamba-based Pedestrian Attribute Recognition}

\author{Xiao Wang, \emph{Member, IEEE}, Weizhe Kong, Jiandong Jin, Shiao Wang, Ruichong Gao, \\ Qingchuan Ma, Chenglong Li, Jin Tang

\thanks{Xiao Wang, Jin Tang are with the School of Computer Science and Technology, Anhui University, Institute of Artificial Intelligence, Hefei Comprehensive National Science Center, Hefei 230601, China. (email: xiaowang@ahu.edu.cn, tangjin@ahu.edu.cn, cheng.zhang@ahu.edu.cn)} 

\thanks{* Corresponding Author: } 
}

\markboth{ IEEE Transactions on Multimedia, 2024 } 
{Shell \MakeLowercase{\textit{et al.}}: Bare Demo of IEEEtran.cls for IEEE Journals}

\maketitle

\begin{abstract}
Current strong pedestrian attribute recognition models are developed based on Transformer networks, which are computationally heavy. Recently proposed models with linear complexity (e.g., Mamba) have garnered significant attention and have achieved a good balance between accuracy and computational cost across a variety of visual tasks. Relevant review articles also suggest that while these models can perform well on some pedestrian attribute recognition datasets, they are generally weaker than the corresponding Transformer models. To further tap into the potential of the novel Mamba architecture for PAR tasks, this paper designs and adapts Mamba into two typical PAR frameworks, i.e., the text-image fusion approach and pure vision Mamba multi-label recognition framework. It is found that interacting with attribute tags as additional input does not always lead to an improvement, specifically, Vim can be enhanced, but VMamba cannot. This paper further designs various hybrid Mamba-Transformer variants and conducts thorough experimental validations. These experimental results indicate that simply enhancing Mamba with a Transformer does not always lead to performance improvements but yields better results under certain settings. We hope this empirical study can further inspire research in Mamba for PAR, and even extend into the domain of multi-label recognition, through the design of these network structures and comprehensive experimentation.
The source code and pre-trained models will be released on \url{https://github.com/Event-AHU/OpenPAR}. 
\end{abstract}

\begin{IEEEkeywords}
Mamba, PAR, Hybrid architecture
\end{IEEEkeywords}

\IEEEpeerreviewmaketitle

\section{Introduction}

\IEEEPARstart{P}{edestrian} Attribute Recognition (PAR), as surveyed in \cite{wang2022PARsurvey}, is a prevalent and significant research area within the computer vision (CV) community. The primary objective of PAR is to discern human attributes based on a variety of descriptive cues, such as \textit{short black hair}, \textit{wearing hats}, \textit{carrying a backpack}, and so forth. This capability is twofold: firstly, it provides a detailed description of pedestrians' appearances and movements, which is essential for a comprehensive understanding of their behavior; secondly, these attributes act as mid-level semantic representations that can significantly enhance the performance of various visual tasks. Such tasks include, but are not limited to, pedestrian detection \cite{9184102}, person re-identification \cite{zhai2024multi, zhu2023attribute}, and others.


Leveraging the power of deep learning, a multitude of sophisticated Pedestrian Attribute Recognition (PAR) models have been introduced, encompassing a range of architectures such as Convolutional Neural Networks (CNNs), Recurrent Neural Networks (RNNs), Graph Neural Networks (GNNs), and Transformers. For instance, the Multi-Task Convolutional Neural Network (MTCNN) \cite{2015mtcnn} employs CNNs to extract features and integrates them in a multi-task learning framework to address the PAR challenge. Joint Representation Learning (JRL) \cite{wang2017JRL} and Graph Representation Learning (GRL) \cite{zhao2018GRL} approach PAR as a sequence prediction problem, utilizing Long Short-Term Memory (LSTM) networks to capture the inter-attribute relationships. More recently, models like PARformer \cite{fan2023parformer} and Visual-Textual Binding (VTB) \cite{cheng2022VTB} have adopted the Transformer architecture for PAR, with VTB notably incorporating textual information to enhance performance.

Despite the success of these existing models in straightforward scenarios, their effectiveness is often constrained in complex environments, such as those with poor lighting or cross-domain variations. Furthermore, the computational expense during training and inference, which scales with $\mathcal{O}(N^2)$ due to the self-attention mechanism inherent in Transformer models, remains a significant challenge.

Given these considerations, it is pertinent to inquire: \textit{How can we devise a novel framework for pedestrian attribute recognition that not only rivals or surpasses the performance of Transformer-based models but also substantially reduces the associated computational costs?} This question drives the pursuit of innovative solutions that balance efficiency with accuracy in the realm of PAR.


The State Space Model (SSM) has recently garnered significant attention within the artificial intelligence community, particularly for its linear complexity (\(\mathcal{O}(N)\)) and impressive performance across various applications. It has been extensively applied to tasks such as visual tracking~\cite{huang2024mambaFETrack}, image and video-based classification~\cite{zhu2024vim, li2024videomamba}, and time series analysis~\cite{wang2024mambaTimeSeries}. The SSM survey~\cite{wang2024SSMsurvey} presents compelling evidence that models like Vim-S~\cite{zhu2024vim}, which leverage SSM for attribute recognition, outperform their ViT-S counterparts on the PA100K~\cite{2017pa100k} dataset, although they may underperform on the PETA~\cite{deng2014peta} dataset. This disparity highlights the need for innovative architectures inspired by Mamba that can consistently enhance performance across different benchmarks while maintaining low computational costs for both training and testing. The development of such frameworks remains a challenging and active area of research, with the potential to significantly advance the field of pedestrian attribute recognition.


\begin{figure*}
    \centering
    \includegraphics[width=0.95\linewidth]{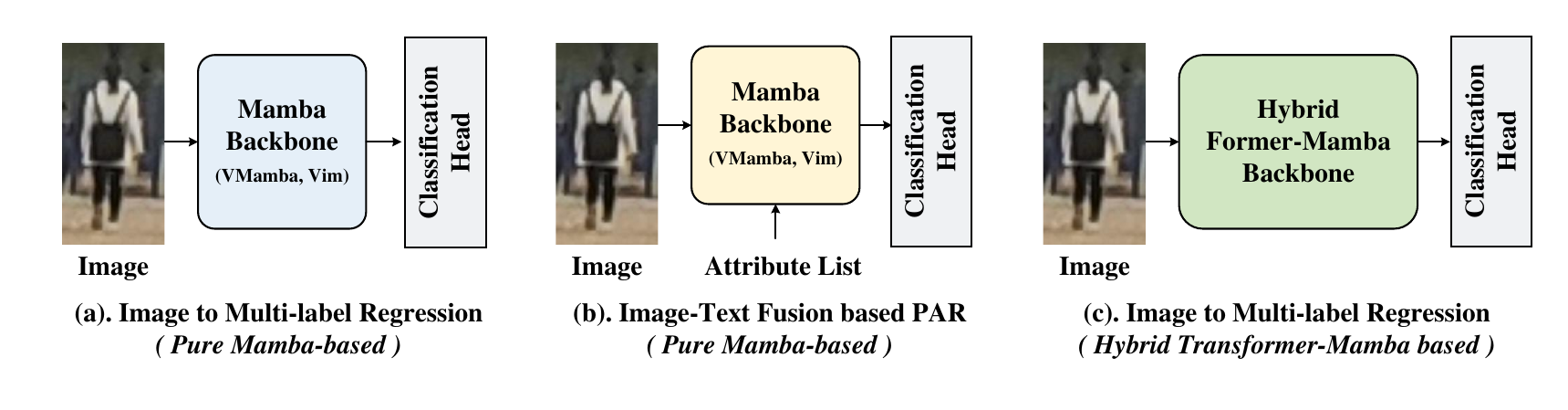}
    \caption{An illustration of our proposed Mamba-based pedestrian attribute recognition framework.}
    \label{fig:firstIMG}
\end{figure*}

Inspired by these works, in this paper, we attempt to conduct extensive experiments to validate the effectiveness of Mamba-based pedestrian attribute recognition. Generally speaking, as illustrated in Fig.~\ref{fig:firstIMG}, we design two representative Mamba-based PAR frameworks, i.e., \textit{image-based multi-label classification PAR framework} and \textit{image-text fusion-based PAR framework}. Our experimental results demonstrate that the pure Mamba-based model performs better than the image-text fusion based model when the VMamba~\cite{liu2024vmamba} is adopted, but the opposite when the Vim~\cite{zhu2024vim} is utilized. 
Furthermore, we consider the design of introducing Transformers to create a more powerful and efficient hybrid Mamba-Transformer backbone architecture for better feature representation extraction, in order to improve the performance of attribute recognition, as illustrated in Fig.~\ref{fig:HybridMambaFormer}. By conducting extensive experiments, we find that when utilizing Mamba for the dense fusion of hierarchical Transformer features, i.e., the Fig.~\ref{fig:HybridMambaFormer} (e), the final results can beat the ViT-B based PAR framework. Also, when distilling from Transformer to Mamba for PAR, i.e., the Fig.~\ref{fig:HybridMambaFormer} (g, h), the performance can also be improved over the Vim-S and ViT-S based version.

To sum up, we conclude the key contributions of this paper with the following three aspects: 

1). We exploit the integration of the recently released Mamba architectures with the PAR task by proposing two representative Mamba-based PAR frameworks, i.e., \textit{the image based multi-label classification framework}, and \textit{image-text fusion based PAR}. Our experimental results indicate that different visual Mamba models are suitable for different PAR frameworks. 

2). We propose several variations of hybrid Transformer-Mamba networks to improve the final attribute recognition performance further and hope it provides comprehensive guidelines for Mamba-based attribute recognition. 

3). We conduct extensive experiments on multiple PAR benchmark datasets, including PA100K~\cite{2017pa100k}, PETA~\cite{deng2014peta}, RAP-V1~\cite{2016rapv1}, RAP-V2~\cite{2019rapv2}, WIDER~\cite{li2016human}, PETA-ZS~\cite{2021Rethinking}, and RAP-ZS~\cite{2021Rethinking}, to demonstrate the effectiveness and efficiency of the proposed Mamba-based PAR frameworks.

\textit{This paper is organized as follows}: 
Firstly, we review the related works on pedestrian attribute recognition and SSM; 
then, we will introduce the pure Mamba-based and Mamba-based vision-language fusion PAR frameworks; 
we also propose different variations of hybrid Mamba-Transformer networks for the PAR.  
After that, we will jump into the experiments, compare them with other state-of-the-art PAR models, and discuss the influence of different settings for Mamba-based PAR. 
We also give some analysis of the model efficiency and visualizations. 
Finally, we conclude this paper and propose possible future works worth to be studied.

\section{Related Works} 
In this section, we will introduce the related works on pedestrian attribute recognition, and state space model. More details can be found in the following surveys~\cite{wang2022PARsurvey, wang2024SSMsurvey, wang2023MMPTMSurvey}.

\subsection{Pedestrian Attribute Recognition}

Pedestrian attribute recognition has experienced remarkable advancements over the years, with a multitude of innovative methods proposed by the research community that have yielded encouraging outcomes. Initially, the primary approach involved the utilization of Convolutional Neural Networks (CNNs) for multi-task training, which facilitated the sharing of parameters across networks to enrich feature extraction. A prime example of this approach is MTCNN~\cite{2015mtcnn}, which embraced multi-task training to extract image features using CNNs and fostered knowledge sharing across various attribute categories. Given the inherent correlations between different attributes, sequential prediction models were subsequently integrated into pedestrian attribute recognition research. JRL~\cite{wang2017JRL} employed Long Short-Term Memory (LSTM) networks to capture contextual relationships both between pedestrians and within attribute sets, effectively framing pedestrian attribute recognition as a sequence prediction challenge. Expanding on the foundation laid by JRL, GRL~\cite{zhao2018GRL} segmented the attribute list into multiple groups and leveraged LSTM to model the spatial and semantic inter-group correlations.

As deep learning networks evolved, attention models started to be incorporated into pedestrian attribute recognition, leading to more nuanced feature extraction. Notably, HydraPlus-Net~\cite{Liu_2017_ICCV} incorporated a multi-directional attention module to distill multi-scale attention features, thereby acquiring a more comprehensive set of pedestrian attributes. The Transformer model, introduced by Vaswani et al.~\cite{NIPS2017_3f5ee243} and widely celebrated in the field of natural language processing, has also been successfully adapted for use in pedestrian attribute recognition. Fan et al.~\cite{fan2023parformer} proposed a purely Transformer-based network for PAR, employing a Transformer backbone for image feature extraction to obtain more discriminative features. Cheng et al.~\cite{cheng2022VTB} introduced VTB, a model that encodes pedestrian images into visual features and attribute annotations into text features using pre-trained text encoders, subsequently employing Transformer for multimodal fusion.

However, the global attention mechanism characteristic of Transformers poses a challenge, as its computational complexity increases quadratically with the length of the input sequence. To address this, Gu et al.~\cite{gu2023mamba} introduced Mamba, a State Space Model designed to optimize sequence processing complexity to a linear scale. Consequently, we advocate for the adoption of a state space model to enhance optimization, thereby improving computational efficiency without compromising the accuracy of pedestrian attribute recognition.

\subsection{State Space Model}

The State Space Model (SSMs)~\cite{gu2021combining,gu2021efficiently,gu2022parameterization,gu2023mamba} has emerged as a prominent mathematical framework for modeling dynamic systems, garnering significant attention for its exceptional capability to manage long sequence data. S4~\cite{gu2021efficiently}, a groundbreaking approach in the application of state space models for long sequence processing, incorporated Hippo~\cite{gu2020hippo} to tackle long-range dependencies in sequence modeling, thereby markedly enhancing model performance. Building on this, S4ND~\cite{nguyen2022s4nd} expanded the spatial state model to encompass multidimensional signals, allowing it to process large-scale visual data as continuous multidimensional signals. This development laid the groundwork for the application of SSMs in multimodal tasks.

Mamba~\cite{gu2023mamba} introduced a novel selective mechanism for information processing, striking a balance between training and inference efficiency while maintaining model effectiveness, and has been shown to surpass the Transformer~\cite{NIPS2017_3f5ee243} in natural language processing tasks. In the wake of Mamba's success, several works have begun to apply its principles to the visual domain, such as Vim~\cite{zhu2024vim} and VMamba~\cite{liu2024vmamba}. These efforts have demonstrated the potential of Mamba in vision-related tasks. Vim introduced a bidirectional Mamba model for image processing, while VMamba proposed a four-direction cross-scan module, which more effectively models two-dimensional image data through a multi-directional sequence order, all while preserving linear computational complexity.

Mamba-2~\cite{dao2024transformers} represents an enhancement over the original Mamba model, introducing the SSD (State Space Duality) framework. This advancement has not only accelerated Mamba's performance by 2-8 times but also achieved superior results.


Our research delves into the potential of Mamba for handling visual-text fusion features and applies Mamba and its derivative modules comprehensively to pedestrian attribute recognition. Through these explorations, we aim to leverage the strengths of Mamba in both visual and textual domains to enhance the accuracy and efficiency of attribute recognition in pedestrians.

\section{Our Proposed Approach} 
In this section, we will first give a brief review of the State Space Model (SSM), then, we will introduce two Mamba-based PAR frameworks, including the \textit{pure image-based multi-label classification} and \textit{image-text fusion based PAR framework}. Further, we design eight variations of hybrid Mamba-Transformer networks for the PAR task. 

\subsection{Preliminary: State Space Model}  
The State Space Model (SSM) originates from the classic Kalman filter~\cite{kalman1960new} algorithm which introduces linear filtering. It converts a one-dimensional sequence into an N-dimensional hidden state for output. The calculation formula is as follows,
\begin{equation}  
\begin{aligned} 
\label{continuous SSMs}
h'(t) = \mathbf{A}h(t) + \mathbf{B}x(t),    \\  
y(t) = \mathbf{C}h(t) +  \mathbf{D}x(t).
\end{aligned}  
\end{equation}
where $x(t) \in \mathbb{R}^{L}$ and $h'(t) \in \mathbb{R}^{N}$ are the input sequence and the derivative of the hidden state respectively. $\mathbf{A} \in \mathbb{R}^{N \times N} $ is the state matrix, $\mathbf{B} \in \mathbb{R}^{N \times L} $ is the input matrix , $\mathbf{C} \in \mathbb{R}^{L \times N} $ is the output matrix, and $\mathbf{D} \in \mathbb{R}^{L \times L}$ is the feed-through matrix. 
The above formula is a continuous-time SSM. In order to facilitate the deep learning algorithm's understanding of the input, we need to discretize the matrixes through some methods, such as zero order hold (ZOH), etc. Specifically, Gu et al.~\cite{gu2021efficiently} propose structured state-space sequence models (S4), which convert continuous parameters into discrete ones by introducing timescale parameter $\Delta$. 
The formula is as follows,
\begin{equation}  
\begin{aligned} 
\label{discretize parameter}
\overline{\mathbf{A}} &= \exp(\Delta\mathbf{A}), \\
\overline{\mathbf{B}} &= (\Delta\mathbf{A})^{-1}(\exp(\Delta\mathbf{A}) - \mathbf{I}) \cdot \Delta\mathbf{B},\\
\overline{\mathbf{C}} &= \mathbf{C}.
\end{aligned}  
\end{equation}
where $\bar{\mathbf{A}}$, $\bar{\mathbf{B}}$, and $\bar{\mathbf{C}}$ are discrete parameters of the system from $\mathbf{A}$, $\mathbf{B}$ and $\mathbf{C}$ respectively. 
Furthermore, the formula can be shown that, 
\begin{equation}  
\begin{aligned} 
\label{discretize SSMs}
{h}_t &=  \bar{\mathbf{A}}{h}_{t-1} + \bar{\mathbf{B}}{x}_t, \\
{y}_t &= \bar{\mathbf{C}}{h}_t.
\end{aligned}  
\end{equation}
where the $\mathbf{D}$ matrix can sometimes be ignored as a residual. Based on the aforementioned models, in the field of computer vision, Vim~\cite{zhu2024vim} and VMamba~\cite{liu2024vmamba} have been proposed and received considerable attention. This paper will explore SSM-based pedestrian attribute recognition frameworks using these two models.

\begin{figure*}
\centering
\includegraphics[width=0.95\linewidth]{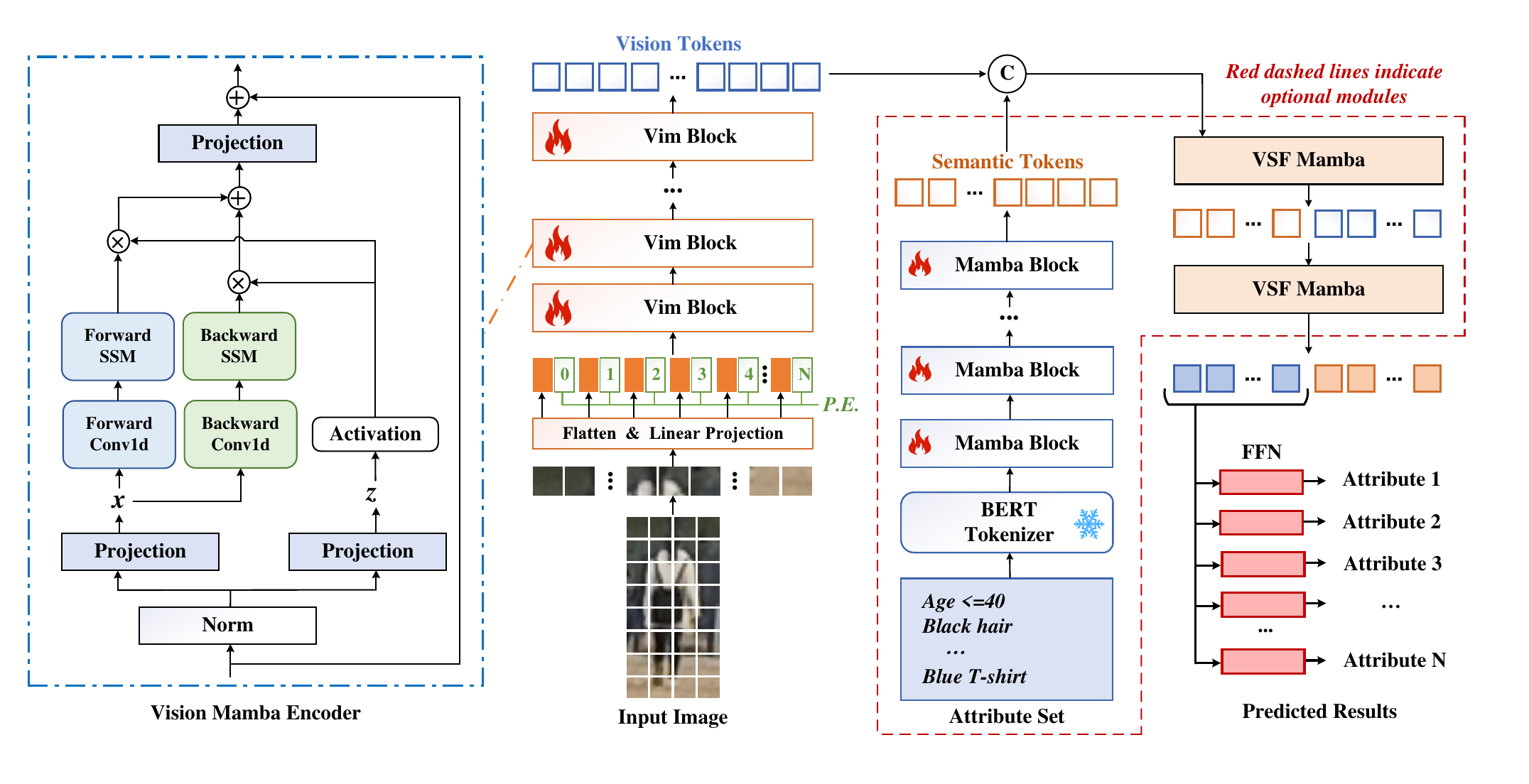}
\caption{An illustration of our proposed vision-language fusion framework based on Mamba for pedestrian attribute recognition.} 
\label{fig:MambaPAR}
\end{figure*}

\subsection{When Mamba Meets PAR} \label{MLS} 

In this section, we mainly focus on two widely used PAR frameworks, i.e., the \textit{image-based multi-label classification framework}~\cite{deepmar,fan2023parformer,2022drformer}, and the \textit{image-text fusion based PAR framework}~\cite{cheng2022VTB,wang2023pedestrianattributerecognitionclip}, as shown in Fig.~\ref{fig:MambaPAR}. Considering that image-based multi-label recognition can be seen as a special case of the image-text integration framework, that is, by removing the attribute label encoding module, this paper will focus on elaborating the PAR process of the image-text integration framework. The details of multi-label recognition will not be further discussed.

Generally speaking, our proposed Mamba-based visual-language fusion PAR architecture consists of three main modules, i.e., the Mamba vision backbone which is composed of a Vim~\cite{zhu2024vim} network, a Mamba-based text encoder for extracting the attribute semantic information, and a visual-semantic fusion (VSF) Mamba module for image-text feature interaction and fusion. Finally, we adopt a classification head for pedestrian attribute recognition. We will introduce these modules in the subsequent paragraphs respectively.

\noindent \textbf{Mamba for Input Representation.~}  
Given a pedestrian image $I$, we need to predict its attributes from an attribute set $A = \{a_1, a_2, ..., a_L\}$, $L$ is the number of person attributes, and the ground truth annotations $Y$ are provided for the supervised learning of our PAR framework. In our framework, we adopt the vision Mamba network Vim~\cite{zhu2024vim} as an example to demonstrate how to encode the given person image into visual features $X_v$. Note that, both the Vim~\cite{zhu2024vim} and VMamba~\cite{liu2024vmamba} can be adopted for the encoding of pedestrian images, as validated in our experimental results. Firstly, we divide the image into equally sized image patches based on a specified patch size, then, these patches are flattened and fed into a linear projection to obtain the patch tokens $X_v \in \mathbb{R}^{P^2 \times D}$. Positional embeddings $P.E.$ are added to each token to better capture the spatial positions of each patch.

Next, we feed the tokens into a stack of Vim~\cite{zhu2024vim} blocks,  $M^V=\{ M_0^V, M_1^V, \dots, M_N^V\}$, where $N$ is the number of Vim blocks, thus, we can obtain the visual tokens $X_v$. More in detail, as shown in Fig.~\ref{fig:MambaPAR} the input tokens are firstly processed using the normalization layer and then transformed into $x$ and $z$ using projection layers. For the $x$, the forward and backward processing branches are adopted for the vision feature learning, and each branch contains both Conv1d and SSM layers. For the $z$, an activation layer $\sigma(\cdot)$ is adopted to enhance the feature and multiply with the output of the forward and backward processing branches respectively. The compute procedure can be simply written as: 
\begin{equation}  
\small 
\begin{aligned} 
\label{Vim}
\sigma(z) * FSSM(Conv1d(x)) + \sigma(z) * BSSM(Conv1d(x)) 
\end{aligned}  
\end{equation}
The output features are added and projected as the output of each Mamba block.

To better help the framework understand what is human attributes, the attribute labels defined on the whole dataset are usually directly integrated into the PAR framework. Given the attribute set $A$, we use the pre-trained BERT~\cite{kenton2019bert} tokenizer to embed the attributes into semantic tokens $X_s$. These semantic tokens are further enhanced by the text Mamba~\cite{gu2023mamba} blocks $M^T=\{ M_0^T, M_1^T, \dots, M_K^T\}$ and the output attribute tokens are fed into vision-language aggregation module which will be introduced in the following paragraphs.

\noindent \textbf{[{Optional}] Mamba for Vision-Language Aggregation.~} 
Previous purely visual PAR methods struggle to effectively correlate attributes with visual features, and the inconsistency of optimization objectives across multiple attribute classifications sharing the same features leads to interference issues. We address these two problems through visual-language interaction. In our visual-language interaction processing, we integrate the extracted visual and textual features, denoted as $\mathcal{X}=[X_v, X_s]$, and input them into the Vision-Semantic Fusion (VSF) Mamba module. This module, consisting of a stack of Mamba blocks, is designed to model the relationship between attributes and visual data across various layers. Ultimately, we employ the textual features $\mathcal{X}_F$ that are aggregated with visual information for classification.

\begin{figure*}
\centering
\includegraphics[width=1\linewidth]{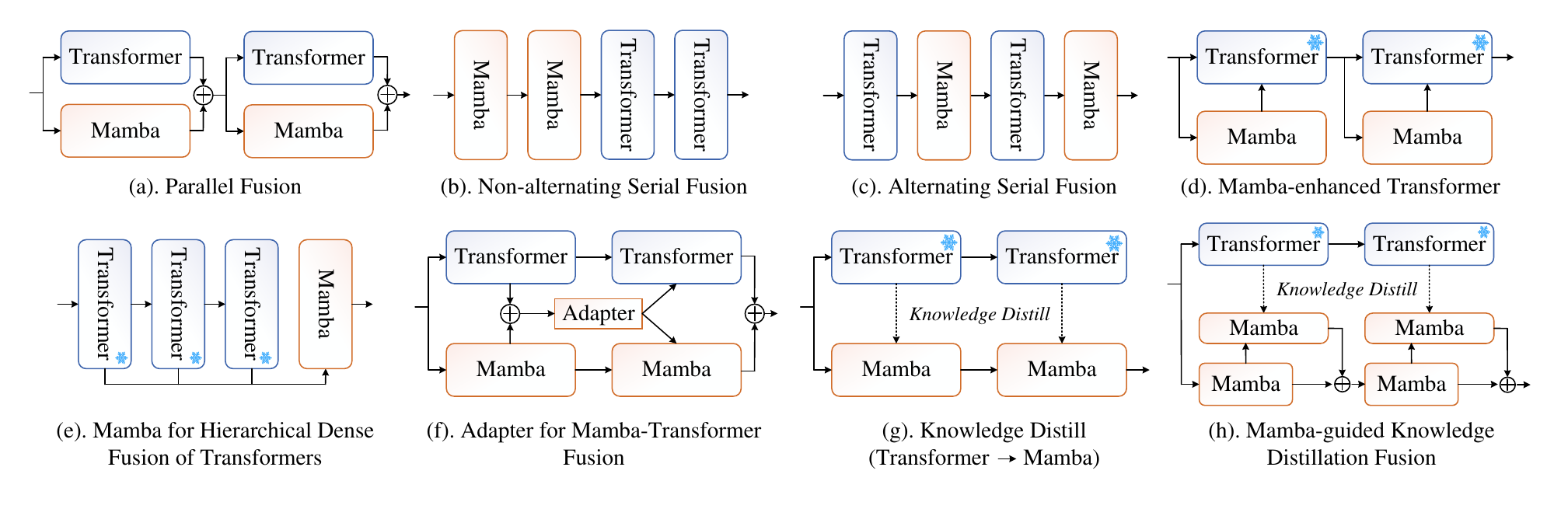}
\caption{An illustration of various hybrid Mamba-Transformer frameworks for PAR.} 
\label{fig:HybridMambaFormer}
\end{figure*}

\noindent \textbf{PAR Classification Head and Loss Function.~} 
By using the Mamba networks to obtain effective feature representations, we input the features to the pedestrian attribute recognition prediction head for multi-label prediction. Specifically, our prediction head is composed of multiple feed-forward networks (FFN). It inputs features of pedestrian attributes to FFNs, outputs the respective category of each input, and completes the final prediction. Specifically, \begin{equation}
P = Sigmoid(FFN(\mathcal{X}))
\end{equation}
where $P=\{p_1, p_2,\dots, p_L \}$, and $L$ is the number of attributes. 
We use the weighted cross-entropy loss as the loss function for our prediction which can be formulated as: 
\begin{equation}
\mathcal{L}= - \sum_{j=1}^L w_j(y_{j}\log(p_{j})+(1-y_{j})\log(1-p_{j}))
\end{equation}
where $w$ is positively correlated with the attribute positive ratio in the training set.

\subsection{Hybrid Mamba-Transformer for PAR}  \label{sec:HybridMamFormer}

In the experiments, we find the MambaPAR still has some performance differences directly with existing transformer-based methods. Therefore, in this section, we exploit the new network architecture to explore Mamba aggregation with the Transformer network for better pedestrian attribute recognition. Specifically speaking, we design eight variations of hybrid Mamba-Transformer networks to validate which one performs better for the PAR task. It should be noted that we do not use Vmamba in the hybrid architecture, but use Vim to mix with Transformer, first because the performance gap between Vim and Transformer is larger, and second because the GPU used in our experiments is a single RTX3090 with 24G memory. It cannot support our hybrid architecture experiment of VMamba and Transformer. Due to the limited space in this paper, we briefly introduced the computation procedures of these hybrid Mamba-Transformer networks for the PAR task. For more details on these models, please check our source code.

\noindent $\bullet$ \textbf{Parallel Fusion (PaFusion)} is shown in Fig.~\ref{fig:HybridMambaFormer} (a), where we connect the corresponding layers in the two models ViT-B and Vim-S in parallel, then add them together and feed into the next layer uniformly. It should be noted that the feature dimension of Vim-S is 384-D, while the feature dimension of ViT-B is 768-D. Thus, linear mapping is performed on the Vim features before feature addition each time. The integrated features are also subjected to reverse linear mapping for input to the next layer of Vim. Since Vim-S contains 24 Mamba layers and ViT-B contains 12 Transformer layers, we connect two Vim layers and one ViT layer in parallel in our practical implementation.

\noindent $\bullet$ \textbf{Non-alternating Serial Fusion (N-ASF)} 
As shown in Fig.~\ref{fig:HybridMambaFormer} (b), we directly connect the Vim-S and ViT-B to build the N-ASF PAR model. After a couple of Vim-S layers, we perform a dimension transformation and feed it into four ViT-B blocks for the feature processing in the next step.

\noindent $\bullet$ \textbf{Alternating Serial Fusion (ASF)} is shown in Fig.~\ref{fig:HybridMambaFormer} (c), where we alternately connect two models of ViT-B and Vim-S layers in serial. There is a linear mapping between the two-layer connections due to the inconsistent number of layers. We connect one layer of ViT to two layers of Vim in series.

\noindent $\bullet$ \textbf{Mamba-enhanced Transformer (MaFormer)} is shown in Fig.~\ref{fig:HybridMambaFormer} (d), where we use Vim-S as a complementary information extractor to integrate the features extracted by Vim and the current layer input before feeding into ViT.

\noindent $\bullet$ \textbf{Mamba for Hierarchical Dense Fusion of Transformer (MaHDFT)} is shown in Fig.~\ref{fig:HybridMambaFormer} (e), where we consider Vim-S as a processor that integrates the outputs of all layers of ViT. We use the ViT-B model trained in advance on the PA100k dataset and then freeze it. Following this, we integrate the outputs of each layer of ViT into four layers of Vim for integration, and then feed the outputs of Vim into the classification head for label prediction. Due to the limited memory of our used RTX3090 GPUs, it is not possible to integrate all tokens directly (2352 total tokens), so each output layer is first passed through a convolutional layer to reduce the number of tokens to 49 (588 total tokens). It should be mentioned here that we retain the classification head of ViT-B on the PA100K dataset for generating the labels corresponding to the final output features of ViT. The classification head used for Vim is recreated, and after obtaining the two sets of label distributions, we average them to obtain the final label.

\noindent $\bullet$ \textbf{Adapter for Mamba-Transformer Fusion (AdaMTF)} is shown in Fig.~\ref{fig:HybridMambaFormer} (f), where we connect Vim and ViT backbone as a whole in parallel, and introduce adapter in the corresponding layer to perform feature interaction, where the adapter is an MLP in our implementation.

\noindent $\bullet$ \textbf{Knowledge Distill (KDTM)} is shown in Fig.~\ref{fig:HybridMambaFormer} (g), where we focus on Vim and use ViT as the teacher network to perform \textit{feature-} or \textit{logit-level} distillation on Vim to enhance the feature representations.

\begin{table*}[!htb]
\center
\small 
\caption{Comparison with state-of-the-art methods on PETA and PA100K datasets. ``V" means Vision branch, ``L" means Language branch. ``-" means this indicator is not available.} 
\label{Comparisononpetadatasets}
\resizebox{\textwidth}{!}{
\begin{tabular}{l|c|ccccc|ccccc}
\hline \toprule [0.5 pt] 
\multicolumn{1}{c|}{\multirow{2}{*}{\textbf{Methods}}} & \multicolumn{1}{c|}{\multirow{2}{*}{\textbf{Backbone}}} & \multicolumn{5}{c|}{\textbf{PETA}} & \multicolumn{5}{c}{\textbf{PA100K}} \\ \cline{3-12} 
\multicolumn{1}{c|}{} &
  \multicolumn{1}{c|}{} &
  \multicolumn{1}{c}{mA} &
  \multicolumn{1}{c}{Acc} & 
  \multicolumn{1}{c}{Prec} &
  \multicolumn{1}{c}{Recall} &
  \multicolumn{1}{c|}{F1} &
  \multicolumn{1}{c}{mA} &
  \multicolumn{1}{c}{Acc} &
  \multicolumn{1}{c}{Prec} &
  \multicolumn{1}{c}{Recall} &
  \multicolumn{1}{c}{F1} \\ \hline
DeepMAR \cite{deepmar} {\scriptsize ACPR15} & CaffeNet & 82.89 & 75.07 & 83.68 & 83.14 & 83.41 & 72.70 & 70.39 & 82.24 & 80.42 & 81.32 \\
HPNet \cite{2017pa100k} {\scriptsize ICCV17} & Inception & 81.77 & 76.13 & 84.92 & 83.24 & 84.07 & 74.21 & 72.19 & 82.97 & 82.09 & 82.53 \\
JRL \cite{wang2017JRL} {\scriptsize ICCV17} & AlexNet & 82.13 & - & 82.55 & 82.12 & 82.02  & - & - & - & - & - \\
GRL \cite{zhao2018GRL} {\scriptsize IJCAI18} & Inception-V3 & 86.70 & - & 84.34 & 88.82 & 86.51  & - & - & - & - & - \\
MsVAA \cite{2018msvaa} {\scriptsize ECCV18} & ResNet101 & 84.59 & 78.56 & 86.79 & 86.12 & 86.46  & - & - & - & - & - \\
RA \cite{zhao2019recurrent} {\scriptsize AAAI19} & Inception-V3 & 86.11 & - & 84.69 & 88.51 & 86.56  & - & - & - & - & - \\
VRKD \cite{2019vrkd} {\scriptsize IJCAI19} & ResNet50 & 84.90 & 80.95 & 88.37 & 87.47 & 87.91 & 77.87 & 78.49 & 88.42 & 86.08 & 87.24 \\
AAP \cite{2019aap} {\scriptsize IJCAI19} & ResNet50 & 86.97 & 79.95 & 87.58 & 87.73 & 87.65 & 80.56 & 78.30 & \textcolor{black}{{89.49}} & 84.36 & 86.85 \\
VAC \cite{2019VAC} {\scriptsize CVPR19} & ResNet50  & - & - & - & - & - & 79.16 & 79.44 & 88.97 & 86.26 & 87.59 \\
ALM \cite{2019alm} {\scriptsize ICCV19} & BN-Inception & 86.30 &  79.52 & 85.65 & 88.09 & 86.85 & 80.68 & 77.08 & 84.24 & 88.84 & 86.46 \\
JLAC \cite{2020JLAC} {\scriptsize AAAI20} & ResNet50 & 86.96 & 80.38 & 87.81 & 87.09 & 87.50 & 82.31 & 79.47 & 87.45 & 87.77 & 87.61 \\
SCRL \cite{wu2020person} {\scriptsize TCSVT20} & ResNet50 & 87.2 & -  & \textcolor{black}{{89.20}} & 87.5 & 88.3 & 80.6 & - & 88.7 & 84.9 & 82.1  \\
SSCNet \cite{2021ssc} {\scriptsize ICCV21} & ResNet50 & 86.52 & 78.95 & 86.02 & 87.12 & 86.99 & 81.87 & 78.89 & 85.98 & 89.10 & 86.87 \\				
IAA-Caps \cite{2022iaacaps} {\scriptsize PR22} & OSNet & 85.27 & 78.04 & 86.08 & 85.80 & 85.64 & 81.94 & 80.31 & 88.36 & 88.01 & 87.80 \\
MCFL \cite{Chen2022MCFL} {\scriptsize NCA22} & ResNet-50 & 86.83 & 78.89 & 84.57 & 88.84 & 86.65 & 81.53 & 77.80 & 85.11 & 88.20 & 86.62 \\
DRFormer \cite{2022drformer} {\scriptsize NC22} & ViT-B/16 & \textcolor{black}{{89.96}} & \textcolor{black}{{81.30}} & 85.68 & \textcolor{black}{{91.08}} & \textcolor{black}{{88.30}} & 82.47 & 80.27 & 87.60 & 88.49 & 88.04 \\
VAC-Combine \cite{guo2022visual} {\scriptsize IJCV22} & ResNet50  & - & - & - & - & - & 82.19 & 80.66 & 88.72 & 88.10 & 88.41 \\
DAFL \cite{jia2022learning} {\scriptsize AAAI22} & ResNet50 & 87.07 & 78.88 & 85.78 & 87.03 & 86.40 & 83.54 & 80.13 & 87.01 & 89.19 & 88.09 \\
CGCN \cite{Fan2022CGCN} {\scriptsize TMM22} & ResNet & 87.08 & 79.30 & 83.97 & 89.38 & 86.59 & - & - & - & - & - \\ 
CAS-SAL-FR \cite{yang2021cascaded} {\scriptsize IJCV22} & ResNet50 & 86.40 & 79.93 & 87.03 & 87.33 & 87.18 & 82.86 & 79.64 & 86.81 & 87.79 & 85.18\\ 
VTB \cite{cheng2022VTB} {\scriptsize TCSVT22} & ViT-B/16 & 85.31 & 79.60 & 86.76 & 87.17 & 86.71 & 83.72 & 80.89 & 87.88 & 89.30 & 88.21 \\
VTB* \cite{cheng2022VTB} {\scriptsize TCSVT22} & ViT-L/14 & 86.34 & 79.59 & 86.66 & 87.82 & 86.97 & \textcolor{black}{{85.30}} & \textcolor{black}{{81.76}} & 87.87 & \textcolor{black}{{90.67}} & \textcolor{black}{{88.86}} \\
PARformer \cite{fan2023parformer} {\scriptsize TCSVT23} & Swin-L & 89.32 & 82.86 & 88.06 & 91.98 & 89.06 & 84.46 & 81.13 & 88.09& 91.67  & 88.52 \\
OAGCN \cite{lu2023oagcn} {\scriptsize TMM23} & Swin-L & 89.91 & 82.95 & 88.26 & 89.10 & 88.68 & 83.74 & 80.38 & 84.55& 90.42  & 87.39 \\
SequencePAR \cite{jin2023sequencepar} {\scriptsize arXiv23} & ViT-L/14 & - & 84.92 & 90.44 & 90.73 & 90.46 & - & 83.94 & 90.38 & 90.23 & 90.10 \\
PromptPAR \cite{wang2023pedestrianattributerecognitionclip} {\scriptsize TCSVT24} & ViT-L/14 & 88.76 & 82.84 & 89.04 & 89.74 & 89.18 & 87.47 & 83.78 & 89.27 & 91.70 & 90.15 \\
SSPNet ~\cite{SHEN2024110194} {\scriptsize PR24} & Swin-S & 88.73 & 82.80 & 88.48 & 90.55 & 89.50 & 83.58 & 80.63 & 87.79 & 89.32 & 88.55  \\
\hline \toprule [0.5 pt] 
MambaPAR (V)&Vim-S &81.45 &74.92 &82.68 & 84.27& 83.15& 79.28 & 76.77 & 85.24 & 86.58  & 85.44 \\
MambaPAR (V+L)&Vim-S &84.25 &76.07 &82.73 & 87.20& 84.56&80.87 &79.33 &86.48 &88.79&87.21  \\	
\rowcolor{gray!25}
MambaPAR (V)&VMamba-B &86.28 &80.54 &87.45 & 87.95& 87.45&83.63 &81.11 &87.59 &89.98&88.40  \\	
\rowcolor{gray!25}
MambaPAR (V+L)&VMamba-B &85.01 &78.47 &85.12 & 87.49& 86.00&81.50 &79.55 &87.54 &88.03&87.35  \\	
\hline \toprule [0.5 pt] 
\end{tabular} }
\end{table*}

\noindent $\bullet$ \textbf{Mamba-guided Knowledge Distillation Fusion (MaKDF)} is shown in Fig.~\ref{fig:HybridMambaFormer} (h), where we introduce a Vim module on top of the direct distillation to perform a distillation transition from the ViT teacher network and then integrate this feature back into the original Vim feature. This avoids the difference between ViT and Vim feature extraction, which brings performance damage to Vim.

\section{Experiments}

\subsection{Datasets and Evaluation Metric}  

In our experiments, seven widely used PAR benchmark datasets are evaluated, including PA100K~\cite{2017pa100k}, PETA~\cite{deng2014peta}, RAP-V1~\cite{2016rapv1}, RAP-V2~\cite{2019rapv2}, WIDER~\cite{li2016human}, PETA-ZS~\cite{2021Rethinking}, RAP-ZS~\cite{2021Rethinking}, and MSP60K~\cite{jin2024pedestrianattributerecognitionnew}. A brief introduction to these datasets is given below: 

\noindent $\bullet$ \textbf{PA100K~\cite{2017pa100k} dataset} is the largest pedestrian attribute recognition dataset, which contains 100,000 pedestrian images and 26 binary attributes. In our experiments, we split them into a training and validation set of 90,000 images, and a test subset of the remaining 10,000 images.

\noindent $\bullet$ \textbf{PETA~\cite{deng2014peta} dataset} contains 19,000 outdoor or indoor pedestrian images and 61 binary attributes. These images are divided into 9500 as the training subset, 1900 as the validation subset, and 7600 as the test subset. In the experiment, we selected 35 pedestrian attributes according to the method of ~\cite{deng2014peta}.

\noindent $\bullet$ \textbf{RAP-V1~\cite{2016rapv1} dataset} contains 41585 pedestrian images and 69 binary attributes, of which 33,268 images are used for training. In the current work, 51 attributes are usually selected for training and evaluation. 

\noindent $\bullet$ \textbf{RAP-V2~\cite{2019rapv2} dataset} has 84,928 pedestrian images and 69 binary attributes, of which 67,943 are used for training. We selected 54 attributes to train and evaluate our model. 

\noindent $\bullet$ \textbf{WIDER~\cite{li2016human} dataset} is divided into a training and validation set of 28,345 images and a test set of 29,179 images with a total of 14 attribute labels. Following the default setup, the train-val set is used for training and the performance on the test set is evaluated. 

\noindent $\bullet$ \textbf{PETA-ZS~\cite{2021Rethinking} dataset} is proposed by Jia et al., based on the PETA  dataset, following a zero-shot protocol. The training, validation, and test sets consist of 11,241 $|$ 3,826 $|$ 3,933 samples respectively. For our experiments, we selected 35 common attributes according to Jia et al.~\cite{2021Rethinking}.

\begin{table*}[!htb]
\center
\setlength{\tabcolsep}{5pt}
\small  
\caption{Comparison with state-of-the-art methods on MSP60K datasets. "V" means Vision branch, "L" means Language branch.} \label{Comparisononmspdatasets}
\begin{tabular}{l|c|ccccc|ccccc}
\hline \toprule [0.5 pt] 
\multirow{2}{*}{Methods} & \multirow{2}{*}{Backbone} & \multicolumn{5}{c|}{Random Split} & \multicolumn{5}{c}{Cross-domain Split} \\ \cline{3-12} 
 &
  \multicolumn{1}{c|}{} &
  \multicolumn{1}{c}{mA} &
  \multicolumn{1}{c}{Acc} & 
  \multicolumn{1}{c}{Prec} &
  \multicolumn{1}{c}{Recall} &
  \multicolumn{1}{c|}{F1} &
  \multicolumn{1}{c}{mA} &
  \multicolumn{1}{c}{Acc} &
  \multicolumn{1}{c}{Prec} &
  \multicolumn{1}{c}{Recall} &
  \multicolumn{1}{c}{F1} \\ \hline
DeepMAR \cite{deepmar} {\scriptsize ACPR15}  &CaffeNet & 70.46  & 72.83 & 84.71 & 81.46 & 83.06 & 54.84 & 44.97 & 63.38  & 58.81 & 61.01  \\
RethinkingPAR \cite{2021Rethinking} {\scriptsize arXiv20} & ResNet-101 & 74.01  & 74.20 & 84.17 & 83.94 & 84.06 & 55.98 & 46.52 & 62.85 & 62.09 & 62.47  \\
SSCNet \cite{2021ssc} {\scriptsize ICCV21}  & ResNet-50 & 69.71  & 69.31  & 79.22  & 82.47  & 80.82 & 52.84 & 40.88  &  56.26 & 58.64 & 57.43 \\
VTB \cite{cheng2022VTB} {\scriptsize TCSVT22} & ViT-B/16 & 76.09 & 75.36 & 83.56 & 86.46 & 84.56 & 58.59 & 49.81 & 65.11 & 66.11 & 65.00 \\
DFDT \cite{ZHENG2023105708} {\scriptsize EAAI22} & Swin-T & 74.19 & 76.35 & 85.03 & 86.35 & 85.69 & 57.85 & 49.97 & 65.34 & 66.18 & 65.76 \\
Zhou et al. \cite{Zhou2023Co} {\scriptsize IJCAI23} & ConvNeXt-B & 73.07  & 68.76 & 78.38 & 82.10 & 80.20 & 54.26  & 41.91 & 56.23 & 60.11 & 58.11  \\
PARformer \cite{fan2023parformer} {\scriptsize TCSVT23} & Swin-L & 76.14 & 76.67 & 84.77 & 86.93 & 85.44 & 57.96 & 50.63 & 62.28 & 71.04 & 65.82 \\
SequencePAR \cite{jin2023sequencepar} {\scriptsize arXiv23} & ViT-L/14 & 71.88 & 71.99 & 83.24 & 82.29 & 82.29 & 57.88 & 50.27 & 65.81 & 65.79 & 65.37 \\
VTB-PLIP \cite{zuo2023plip} {\scriptsize arXiv23} & ViT-B/16  &73.90  &73.16  &82.01  &84.82  &82.93  &56.30  &46.77  &61.20  &64.47  &62.18  \\
Rethink-PLIP \cite{zuo2023plip} {\scriptsize arXiv23} & ResNet-101  &69.44  &68.90  &79.82  &81.15  &80.48  &57.18  &46.98  &63.57  &62.16  &62.86  \\
PromptPAR \cite{wang2023pedestrianattributerecognitionclip} {\scriptsize TCSVT24} & Vit-L/14 & 78.81 & 76.53 & 84.40 & 87.15 & 85.35 & 63.24 & 53.62 & 66.15 &71.84 & 68.32 \\
SSPNet ~\cite{SHEN2024110194} {\scriptsize PR24} & Swin-S & 74.03  & 74.10  & 84.01  & 84.02  & 84.02  & 56.15  & 46.75  & 62.44  & 63.07  & 62.75  \\
\hline 
MambaPAR (V+L) & Vim-S & 73.85  & 73.64 & 83.19 & 84.29 & 83.28 & 56.75 & 47.34 & 61.92 & 64.98 & 62.80 \\ 
MambaPAR (V) & VMamba-B  & 74.82  & 74.68  & 82.32  & 87.08  & 84.18 & 58.26 & 49.76 & 62.79 & 68.89 & 65.03\\
MaHDFT & Vim-S  & 74.08  & 74.40  & 82.82  & 86.41  & 83.93 & 58.67 & 50.65 & 62.39 & 71.13 & 65.85\\
\hline \toprule [0.5 pt] 
\end{tabular}
\end{table*}

\noindent $\bullet$ \textbf{RAP-ZS~\cite{2021Rethinking} dataset} is developed based on the RAPv2 dataset, where the training, validation, and testing sets contain 17062, 4628, and 4928 pedestrian images, respectively. There is no shared personal identity between training and inference data. In the experiment, we selected 53 attributes for evaluation following Jia et al.~\cite{2021Rethinking}.

\noindent $\bullet$ \textbf{MSP60K~\cite{jin2024pedestrianattributerecognitionnew} dataset} is a newly released pedestrian attribute recognition dataset, which contains 60122 pedestrian images and 57 binary attributes. In the experiment, according to the setting of MSP60K, 30,298 images are used as the training set, 6002 images are used as the validation set, and 23,822 images are used as the test set in the random partition. In the cross-domain partition, 34128 images from five scenes (construction site, market, kitchen, school, ski resort) are used for training, and 24994 images from three scenes (Outdoors1, Outdoors2, Outdoors3) are used for testing.

To evaluate our proposed Mamba-based pedestrian attribute recognition algorithm and other SOTA PAR models, we adopt the following evaluation metrics, including mA, Accuracy, Precision, Recall, and F1-measure. 
For the mA can be expressed as: 
\begin{equation*} 
    \textit{mA} = \frac{\sum_{i=1}^{C} AP_{i}  }{C}
\end{equation*}
where $AP_{i}$ is the area under the precision-recall curve for the $attribute_{i}$, and C is the total number of attributes. 
The Accuracy can be expressed as:
\begin{equation*}
\textit{Accuracy} = \tfrac{TP\:+\:TN}{TP\:+\:TN\:+\:FP +\:FN}
\end{equation*}
where TP is a positive sample predicted correctly, TN is a negative sample predicted correctly, FP is a negative sample predicted incorrectly, and FN is a positive sample predicted incorrectly. The Precision, Recall, and F1-measure can be expressed as: 
\begin{equation*}
\textit{Precision} = \tfrac{TP}{TP\:+\:FP},  \quad
\textit{Recall} = \tfrac{TP}{TP\:+\:FN},  
\end{equation*}
\begin{equation*}
\textit{F1} = \tfrac{2 \times Precision \times Recall}{Precision\:+\:Recall} 
\end{equation*}

\subsection{Implementation Details}  
In our experiments, both VMamba-B and Vim-S versions are used as visual feature extractors. VMamba-B is a vision Mamba network in four stages with 2, 2, 27, and 2 layers in each stage, and the feature dimension is 768-D. Vim-S is a 24-layer vision Mamba model with 384 feature dimensions. For the text modality, the original Mamba model Mamba-130M, with a feature size of 768 is used. Note that, random cropping and flipping operations are employed for data augmentation. 

For the detailed parameters, we train the model for 100 epochs using the Adam~\cite{adam2015} optimizer. When using VMamba-B, we set the learning rate to 8e-5 and the batch size to 16. When using Vim-S, we set the learning to 2.5e-4 and the batch size to 64. Our source code is implemented using Python and the deep learning framework PyTorch~\cite{paszke2019pytorch}. The experiment were conducted on the server using a GPU RTX3090s. More details can be found in our source code.

\subsection{Comparison on Public PAR Benchmarks} 
In this section, we report the recognition results on seven datasets and compare them with existing state-of-the-art pedestrian attribute recognition algorithms. Note that the results for VTB* were obtained by replacing VTB's backbone network with ViT-L/14. Visualization of attribute predictions of some person images is given in Fig.~\ref{fig:PARresults_VIS}.

\begin{table*}
\center
\small
\caption{Comparison with state-of-the-art methods on RAPv1 and RAPv2 datasets. ``V" means Vision branch. ``-" means this indicator is not available.} \label{Comparisononrapdatasets} 
\resizebox{\textwidth}{!}{
\begin{tabular}{l|c|ccccc|ccccc}
\hline \toprule [0.5 pt]
\multicolumn{1}{c|}{\multirow{2}{*}{\textbf{Methods}}} & \multicolumn{1}{c|}{\multirow{2}{*}{\textbf{Backbone}}} & \multicolumn{5}{c|}{\textbf{RAPv1}} & \multicolumn{5}{c}{\textbf{RAPv2}} \\ \cline{3-12} 
\multicolumn{1}{c|}{} &   \multicolumn{1}{c|}{} &  \multicolumn{1}{c}{mA} &  \multicolumn{1}{c}{Acc} &  \multicolumn{1}{c}{Prec} &  \multicolumn{1}{c}{Recall} &  \multicolumn{1}{c|}{F1} &  \multicolumn{1}{c}{mA} &  \multicolumn{1}{c}{Acc} &
  \multicolumn{1}{c}{Prec} &  \multicolumn{1}{c}{Recall} &  \multicolumn{1}{c}{F1} \\ 
\hline
DeepMAR \cite{deepmar} {\scriptsize ACPR15} & CaffeNet & 73.79 & 62.02 & 74.92 & 76.21 & 75.56 & - & - & - & - & - \\
HPNet \cite{2017pa100k} {\scriptsize ICCV17} & Inception & 76.12 & 65.39 & 77.33 & 78.79 & 78.05 & - & - & - & - & - \\
JRL \cite{wang2017JRL} {\scriptsize ICCV17} & AlexNet & 74.74 & - & 75.08 & 74.96 & 74.62 & - & - & - & - & - \\
GRL \cite{zhao2018GRL} {\scriptsize IJCAI18} & Inception-V3 & 81.20 & - & 77.70 & 80.90 & 79.29 & - & - & - & - & - \\
MsVAA \cite{2018msvaa} {\scriptsize ECCV18} & ResNet101 & - & - & - & - & - & 78.34 & 65.57 & \textcolor{black}{{77.37}} & 79.17 & 78.26  \\
RA \cite{zhao2019recurrent} {\scriptsize AAAI19} & Inception-V3 & 81.16 & - & 79.45 & 79.23 & 79.34 & - & - & - & - & - \\
VRKD \cite{2019vrkd} {\scriptsize IJCAI19} & ResNet50 & 78.30 & 69.79 & \textcolor{black}{{82.13}} & 80.35 & 81.23 & - & - & - & - & - \\
AAP \cite{2019aap} {\scriptsize IJCAI19} & ResNet50 & 81.42 & 68.37 & 81.04 & 80.27 & 80.65 & - & - & - & - & - \\
VAC \cite{2019VAC} {\scriptsize CVPR19} & ResNet50 & - & - & - & - & - & 79.23 & 64.51 & 75.77 & 79.43 & 77.10 \\
ALM \cite{2019alm} {\scriptsize ICCV19}  & BN-Inception & 81.87 & 68.17 & 74.71 & 86.48 & 80.16 & 79.79 & 64.79 & 73.93 & 82.03 & 77.77 \\
JLAC \cite{2020JLAC} {\scriptsize AAAI20} & ResNet50 & 83.69 & 69.15 & 79.31 & 82.40 & 80.82 & 79.23 & 64.42 & 75.69 & 79.18 & 77.40 \\
SSCNet \cite{2021ssc} {\scriptsize ICCV21}  & ResNet50 & 82.77 & 68.37 & 75.05 & 87.49 & 80.43 & - & - & - & - & - \\
IAA-Caps \cite{2022iaacaps} {\scriptsize PR22} & OSNet & 81.72 & 68.47 & 79.56 & 82.06 & 80.37 & - & - & - & - & - \\
MCFL \cite{Chen2022MCFL} {\scriptsize NCA22} & ResNet50 & 84.04 & 67.28 & 73.44 & \textcolor{black}{{87.75}} & 79.96 & - & - & - & - & - \\
DRFormer \cite{2022drformer} {\scriptsize NC22} & ViT-B/16 & 81.81 & \textcolor{black}{{70.60}} & 80.12 & 82.77 & 81.42 & - & - & - & - & - \\
VAC-Combine \cite{guo2022visual} {\scriptsize IJCV22} & ResNet50 & 81.30 & 70.12 & \textcolor{black}{{81.56}} & 81.51 & \textcolor{black}{{81.54}} & - & - & - & - & - \\
DAFL \cite{jia2022learning} {\scriptsize AAAI22} & ResNet50 & 83.72 & 68.18 & 77.41 & 83.39 & 80.29 & 81.04 & 66.70 & 76.39 & 82.07 & 79.13 \\
CGCN \cite{Fan2022CGCN} {\scriptsize TMM22} & ResNet50 & \textcolor{black}{{84.70}} & 54.40 & 60.03 & 83.68 & 70.49 & - & - & - & - & -  \\ 
CAS-SAL-FR \cite{yang2021cascaded} {\scriptsize IJCV22} & ResNet50 & 84.18 & 68.59 & 77.56 & 83.81 & 80.56 & - & - & - & - & - \\ 
VTB \cite{cheng2022VTB} {\scriptsize TCSVT22} & ViT-B/16 & 82.67 & 69.44 & 78.28 & 84.39 & 80.84 & 81.34 & 67.48 & 76.41 & 83.32 & 79.35 \\
VTB* \cite{cheng2022VTB} {\scriptsize TCSVT22}  & ViT-L/14 & 83.69 & 69.78 & 78.09 & 85.21 & 81.10 & \textcolor{black}{{81.36}} & \textcolor{black}{{67.58}} & 76.19 & \textcolor{black}{{84.00}} & \textcolor{black}{{79.52}} \\	
PARformer \cite{fan2023parformer} {\scriptsize TCSVT23} & Swin-L & 84.13 & 69.94 & 79.63 & \textcolor{black}{{88.19}} & 81.35 & - & - & -& -  & - \\
OAGCN \cite{lu2023oagcn} {\scriptsize TMM23} & Swin-L & 87.83 & 69.32 & 78.32 & 87.29 & 82.56 & - & - & -& -  & - \\
SequencePAR \cite{jin2023sequencepar} {\scriptsize arXiv23} & ViT-L/14 & - & 71.47 & 82.40 & 82.09 & 82.05 & - & 70.14 & 81.37 & 81.22 & 81.10 \\
PromptPAR \cite{wang2023pedestrianattributerecognitionclip} {\scriptsize TCSVT24}  & ViT-L/14 & \textcolor{black}{{85.45}} & \textcolor{black}{{71.61}} & 79.64 & 86.05 & \textcolor{black}{{82.38}} 
& \textcolor{black}{{83.14}} & \textcolor{black}{{69.62}} & \textcolor{black}{{77.42}} & \textcolor{black}{{85.73}} & \textcolor{black}{{81.00}} \\

SSPNet ~\cite{SHEN2024110194} {\scriptsize PR24} & Swin-S & 83.24 & 70.21 & 80.14 & 82.90 & 81.50 & - & - & - & - & -  \\
\hline 
MambaPAR (V) &VMamba-B & 83.12&69.47 & 78.03&84.81 &80.88 &81.91 & 68.08&76.41 &84.38 & 79.83\\
\hline \toprule [0.5 pt] 
\end{tabular} } 
\end{table*}

\noindent
\textbf{Results on PETA dataset. } 
As shown in Table \ref{Comparisononpetadatasets}, the evaluated MambaPAR model performs well on most evaluation metrics, with different branches and backbone networks showing varying results. Specifically, based on VMamba-B, the MambaPAR(V) achieved mA, Acc, Prec, Recall, and F1 metrics of 86.28, 80.54, 87.45, 87.95, and 87.45, respectively; the MambaPAR(V+L) achieved 85.01, 78.47, 85.12, 87.49, and 86.00. Based on Vim-S, the MambaPAR(V) achieved 81.45, 74.92, 82.68, 84.27, and 83.15; the MambaPAR(V+L) achieved 84.25, 76.07, 82.73, 87.20, and 84.56. 
Compared to the Transformer-based pedestrian attribute recognition models, our VMamba visual branch model surpassed VTB (ViT-B/16, 85.31/79.60/86.76/87.17/86.71) and VTB* (ViT-L/14, 86.34/79.59/86.66/87.82/86.97) in the mA, Acc, Prec, Recall, and F1 metrics. Therefore, we can conclude that our model achieves competitive results on the PETA dataset, verifying the feasibility of the Mamba model for pedestrian attribute recognition.

\noindent
\textbf{Results on PA100K dataset. }
As shown in Table \ref{Comparisononpetadatasets}, our MambaPAR model also performs well on the PA100K dataset. Based on VMamba-B, the MambaPAR(V) achieved mA, Acc, Prec, Recall, and F1 metrics of 83.63, 81.11, 87.59, 89.9, and 88.40, respectively; the MambaPAR(V+L) achieved 81.50, 79.55, 87.54, 88.03, and 87.35. Based on Vim-S, the MambaPAR(V+L) achieved 79.28, 76.77, 85.24, 86.58, and 85.44; the MambaPAR(V) achieved 80.87, 79.33, 86.48, 88.79, and 87.21. The VMamba-B visual branch model's performance is close to that of VTB* (85.3/81.76/87.87/90.67/88.86). These experimental results indicate that the exploited Mamba-based PAR frameworks we explored have achieved preliminary success. 

Interestingly, in the experimental results of PETA and PA100K we can find that the attribute labels are useful for the Vim-S, but not for the VMamba-B. This may be because the textual representation does not match the visual representation, and the direct visual textual feature interaction using Mamba blocks at the end instead limits the representation ability of VMamba. The representation ability of Vim is weak, and the use of Mamba for feature interaction brings effective improvement.

\noindent
\textbf{Results on MSP60K dataset. }
As shown in Table \ref{Comparisononmspdatasets}, in the random partition, our MambaPAR(V) based on VMamba-B achieved mA, Acc, Prec, Recall, and F1 metrics of 74.82, 74.68, 82.32, 86.41, and 84.18, respectively. This goes beyond SSPNet based on Swin-S. The MambaPAR(V+L) based on Vim-S achieved mA, Acc, Prec, Recall, and F1 metrics of 73.85, 73.64, 83.19, 84.29, and 83.28, respectively. The hybrid architecture(MaHDFT) based on Vim-S achieved mA, Acc, Prec, Recall, and F1 metrics of 74.08, 74.40, 82.82, 86.41, and 83.93, respectively. In the cross-domain partition, the MambaPAR(V) based on VMamba-B achieved mA, Acc, Prec, Recall, and F1 metrics of 58.26, 49.76, 62.79, 68.89, and 65.03, respectively. The MambaPAR(V+L) based on Vim-S achieved mA, Acc, Prec, Recall, and F1 metrics of 56.75, 47.34, 61.92, 64.98, and 62.80, respectively. The hybrid architecture(MaHDFT) based on Vim-S achieved mA, Acc, Prec, Recall, and F1 metrics of 58.67, 50.65, 62.39, 71.13, and 65.85, respectively. As can be seen from the results, VMamba and the hybrid architecture model has better generalization. 

\textit{From the experimental results reported in the PETA, PA100K, MSP60K datasets, we can find that the VMamba-B achieves the best results over other Mamba-based PAR models. Therefore, we only report the results of pure VMamba-B based PAR framework for the rest of the datasets. }

\begin{table*}[!htb]
\center
\small  
\caption{Comparison with state-of-the-art methods on PETA-ZS and RAP-ZS datasets. "V" means Vision branch.} \label{PARZSresults} 
\resizebox{\textwidth}{!}{
\begin{tabular}{l|c|ccccc|ccccc}
\hline \toprule [0.5 pt]
\multicolumn{1}{c|}{\multirow{2}{*}{\textbf{Methods}}} & \multicolumn{1}{c|}{\multirow{2}{*}{\textbf{Backbone}}} & \multicolumn{5}{c|}{\textbf{PETA-ZS}} & \multicolumn{5}{c}{\textbf{RAP-ZS}} \\ \cline{3-12} 
\multicolumn{1}{c|}{} &  \multicolumn{1}{c|}{} &  \multicolumn{1}{c}{mA} &  \multicolumn{1}{c}{Acc} &  \multicolumn{1}{c}{Prec} &  \multicolumn{1}{c}{Recall} &  \multicolumn{1}{c|}{F1} &\multicolumn{1}{c}{mA} &  \multicolumn{1}{c}{Acc} &  \multicolumn{1}{c}{Prec} &  \multicolumn{1}{c}{Recall} &  \multicolumn{1}{c}{F1} 
\\ 
\hline
MsVAA \cite{2018msvaa} {\scriptsize ECCV18} & ResNet101 & 71.53 & 58.67 & 74.65 & 69.42 & 71.94 & 72.04 & 62.13 & 75.67 & 75.81 & 75.74\\
VAC \cite{2019VAC} {\scriptsize CVPR19}  & ResNet50 & 71.91 & 57.72 & 72.05 & 70.64 & 70.90 & 73.70 & 63.25 & 76.23 & 76.97 & 76.12 \\
ALM \cite{2019alm} {\scriptsize ICCV19} & BN-Inception & 73.01 & 57.78 & 69.50 & 73.69 & 71.53  & 74.28 & 63.22 & 72.96 & 80.73 & 76.65 \\
JLAC \cite{2020JLAC} {\scriptsize AAAI20}  & ResNet50 & 73.60 & 58.66 & 71.70 & 72.41 & 72.05 & 76.38 & 62.58 & 73.14 & 79.20 & 76.05 \\
Jia et al. \cite{2021Rethinking} {\scriptsize arXiv21} & ResNet50 & 71.62 & 58.19 & 73.09 & 70.33 & 71.68 & 72.32 & 63.61 & \textcolor{black}{{76.88}} & 76.62 & 76.75 \\
MCFL \cite{Chen2022MCFL} {\scriptsize NCA22} & ResNet50 & 72.91 & 57.04 & 68.47 & 74.35 & 71.29 & 74.37 & 63.37 & 71.21 & 83.86 & 77.02 \\
VTB \cite{cheng2022VTB} {\scriptsize TCSVT22} & ViT-B/16    & 75.13 & 60.50 & 73.29 & 74.40 & 73.38 & 75.76 & 64.73 & 74.93 & 80.85 & 77.35 \\
VTB* \cite{cheng2022VTB} {\scriptsize TCSVT22} & ViT-L/14    & \textcolor{black}{{77.18}} & \textcolor{black}{{63.12}} & \textcolor{black}{{74.77}} & \textcolor{black}{{77.24}} & \textcolor{black}{{75.50}} & \textcolor{black}{{79.17}} & \textcolor{black}{{68.34}} &76.81 & \textcolor{black}{{84.51}} & \textcolor{black}{{80.07}} \\
SequencePAR \cite{jin2023sequencepar} {\scriptsize arXiv23}  & ViT-L/14 & - & 66.70 & 78.75 & 78.52 & 78.40 & - & 70.28 & 82.13 & 80.55 & 81.14 \\
PromptPAR \cite{wang2023pedestrianattributerecognitionclip} {\scriptsize TCSVT24}  & ViT-L/14                       & \textcolor{black}{{80.08}} & \textcolor{black}{{66.02}} & \textcolor{black}{{76.53}} & \textcolor{black}{{80.49}} & \textcolor{black}{{77.77}}         & \textcolor{black}{{80.43}} & \textcolor{black}{{70.39}} & \textcolor{black}{{78.48}} & \textcolor{black}{{85.57}} & \textcolor{black}{{81.52}}   \\

\hline 
MambaPAR (V) &VMamba-B & 74.43&60.56 & 73.12&74.51 &73.37 &77.34 & 66.64&75.37 &83.57 & 78.86\\	
\hline \toprule [0.5 pt] 
\end{tabular} } 
\end{table*}

\noindent
\textbf{Results on RAP-V1 dataset.}
As shown in Table \ref{Comparisononrapdatasets}, our VMamba-B visual branch model achieved mA, Acc, Prec, Recall, and F1 metrics of 83.12, 69.47, 78.03, 84.81, and 80.88, respectively. In comparison, the strong baseline VTB* achieved 83.69, 69.78, 78.09, 85.21, and 81.10 on these metrics, and our results are very close to theirs. Meanwhile, our model's performance is also close to the DRFormer~\cite{2022drformer}, which is also developed based on Transformer. This fully demonstrates that the performance of our Mamba-based pedestrian attribute recognition model is comparable to or even surpasses Transformer-based models.

\noindent
\textbf{Results on RAP-V2 dataset.}
As shown in Table \ref{Comparisononrapdatasets}, our VMamba-B visual branch model achieved mA, Acc, Prec, Recall, and F1 metrics of 81.91, 68.08, 76.41, 84.38, and 79.83, respectively, outperforming the Transformer-based VTB* (81.36/67.58/76.19/84.00/79.52). This further proves the effectiveness of our proposed pedestrian attribute recognition model.

\noindent
\textbf{Results on WIDER dataset.}
As shown in Table~\ref{Comparisononwiderdatasets}, our model achieved mA of 89.38, surpassing VTB (ViT-B/16, 88.2), which further validates the feasibility of our model. Compared with the pre-trained vision-language CLIP~\cite{radford2021CLIP} model based PAR, e.g., the VTB* and PromptPAR, the VMamba-based model is still inferior to these models. In our future works, we will consider knowledge distillation strategies to further augment the performance of VMamba-based PAR.

\begin{table}
\center
\small  
\caption{Comparison with state-of-the-art methods on WIDER datasets. ``V" means Vision branch.} \label{Comparisononwiderdatasets}  
\resizebox{0.45\textwidth}{!}{
\begin{tabular}{l|c|c}
\hline \toprule [0.5 pt]
\multicolumn{1}{c|}{\multirow{1}{*}{\textbf{Methods}}} & \multicolumn{1}{c|}{\multirow{1}{*}{\textbf{Backbone}}} & \multicolumn{1}{c}{\textbf{mA}} \\
\hline
R*CNN \cite{2015rcnn} {\scriptsize ICCV15} & VGG16 & 80.5\\
DHC \cite{li2016wider} {\scriptsize ECCV16} & VGG16 & 81.3\\
SRN \cite{2017srn} {\scriptsize CVPR17} & ResNet101 & 86.2\\
DIAA \cite{2018msvaa} {\scriptsize ECCV18} & ResNet101 & 86.4\\
Da-HAR \cite{2019dahar} {\scriptsize AAAI20} & ResNet101 & 87.3\\
VAC-Combine \cite{guo2022visual} {\scriptsize IJCV22} & ResNet50 & 88.4\\
VTB \cite{cheng2022VTB} {\scriptsize TCSVT22} & ViT-B/16 & 88.2\\
VTB* \cite{cheng2022VTB} {\scriptsize TCSVT22} & ViT-L/14 & \textcolor{black}{{91.2}}\\
PromptPAR \cite{wang2023pedestrianattributerecognitionclip} {\scriptsize TCSVT24} & ViT-L/14 &  \textcolor{black}{{92.0}}\\  
\hline 
MambaPAR (V) & VMamba-B & 89.4\\ 
\hline \toprule [0.5 pt] 
\end{tabular} } 
\end{table}

\noindent
\textbf{Results on PETA-ZS dataset.} 
In addition to standard evaluations, we also evaluated the model's performance on datasets using a zero-shot setting. According to the results reported in Table \ref{PARZSresults}, our model achieved mA, Acc, Prec, Recall, and F1 metrics of 74.43, 60.56, 73.12, 74.51, and 73.37, respectively, which are close to or even better than VTB (ViT-B/16, 75.13/60.50/73.29/74.40/73.38) on certain metrics.

\noindent
\textbf{Results on RAP-ZS dataset.} 
As shown in Table \ref{PARZSresults}, the experimental results on the RAP-ZS dataset show that VTB based on ViT-B/16 achieved mA, Acc, Prec, Recall, and F1 metrics of 75.76, 64.73, 74.93, 80.85, and 77.35, respectively, while our experimental results were more excellent, reaching 77.34, 66.64, 75.37, 83.57, and 78.86, respectively. These results indicate that our model also performs excellently in zero-shot settings, and we believe the key reason is the high adaptability of the Mamba model to pedestrian attribute recognition.

\subsection{Ablation Study}

\noindent
\textbf{Effects of Vision-Semantic Fusion for PAR.} 
As shown in Table~\ref{Comparisononpetadatasets}, we conducted ablation experiments on the PA100K and PETA datasets for the visual semantic fusion (VSF) module. The effectiveness of this module has been validated by integrating with the Transformer networks, but not with the Mamba networks. Thus, we exploit two vision Mamba models, i.e., the VMamba-B and Vim-S, to check whether VSF still works for the Mamba-based PAR framework. It is easy to find that the pure Vim-S based model achieves (PA100K) 81.45, 74.92, 82.68, 84.27, 83.15 and (PETA) 79.28, 76.77, 85.24, 86.58, 85.44 in the five indicators of mA, Acc, Prec, Recall, and F1, respectively. After introducing the VSF module highlighted in the red dashed rectangle, the results reach 80.87, 79.33, 86.48, 88.79, 87.21 on the PA100K dataset and 84.25, 76.07, 82.73, 87.20, 84.56 on the PETA dataset, respectively, which proves the effectiveness of the semantic labels for the Vim-based PAR framework. However, we also find that similar conclusions do not hold for the VMamba-B based PAR framework, that is to say, the performance drops when adopting the semantic labels of all attributes.

\begin{figure*}
    \centering
    \includegraphics[width=1\textwidth]{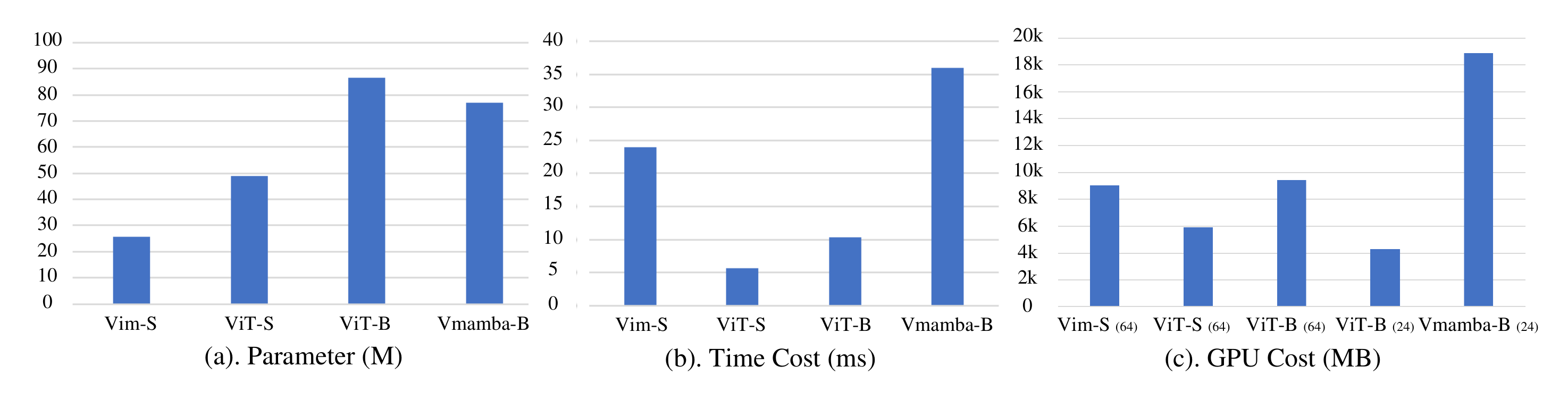}
    \caption{Comparison on (a) model parameters, (b) Time cost, and (c) GPU cost. }
    \label{fig:enter-label}
\end{figure*}

\noindent
\textbf{Transformer \textit{vs} Mamba for PAR.}
As shown in Table \ref{transVsMamba}, we compare the performance of ViT-S, Vim-S, ViT-B, and VMamba-B on the PA100K dataset using only the visual branch. One can find that the vision Mamba-based PAR framework achieves comparable or even better performance than vision Transformer based models at the same size. These comparisons demonstrate that it is still a promising research direction to adopt the Mamba for pedestrian attribute recognition.

\begin{table}
\center
\small  
\caption{Results of different backbones on PA100k dataset. (Only Vision branch) } 
\label{transVsMamba}  
\resizebox{0.45\textwidth}{!}{
\begin{tabular}{l|ccccc}
\hline \toprule [0.5 pt]
\multicolumn{1}{c|}{\multirow{1}{*}{\textbf{Methods}}} & \multicolumn{1}{c}{\textbf{mA}} &  \multicolumn{1}{c}{\textbf{Acc}} &  \multicolumn{1}{c}{\textbf{Prec}} &  \multicolumn{1}{c}{\textbf{Recall}} &  \multicolumn{1}{c}{\textbf{F1}}  \\ 
\hline 
ViT-S & 78.86 & 77.05 & 85.83 & 86.53  & 85.67 \\
Vim-S & 79.28 & 76.77 & 85.24 & 86.58  & 85.44 \\
ViT-B & 83.10 & 80.61 & 87.75 & 89.12  & 88.06 \\
VMamba-B &83.63 &81.11 &87.59 &89.98&88.40  \\
\hline \toprule [0.5 pt] 
\end{tabular} } 
\end{table}

\subsection{Efficiency Analysis}

\noindent \textbf{Analysis of Testing Time. } 
As shown in Fig.~\ref{fig:enter-label}(a), we test the running time of Vim-S, VMamba-B, ViT-S, and ViT-B to check their efficiency. Obviously, the Mamba-based models are slower than the ViT-based models. Also, the VMamba-B needs more time than the Vim-S due to larger parameters. Thus, research on further improving the efficiency of Vision Mamba is needed.

\noindent \textbf{Analysis on GPU Memory Cost. } 
As shown in Fig.~\ref{fig:enter-label}(c), we compared the memory usage of different PAR models on various datasets. It is easy to find that the VMamba significantly costs more GPU memories (about $\times2$ times) than the Vim-S, ViT-S, and ViT-B. Interestingly, we can also see that the ViT-B performs the best with batch size 24 or 64. These experimental results indicate that Mamba does not gain an advantage regarding memory usage on pedestrian attribute recognition tasks with limited image resolutions.

\noindent \textbf{Analysis of Parameters. }  
As illustrated in Fig.~\ref{fig:enter-label}(b), the Vim-S model exhibits a notable reduction in parameters compared to the ViT-S and ViT-B models, with no appreciable decline in performance. Additionally, the VMamba-B model displays a more compact parameter count than the ViT-B model, while maintaining comparable performance. It has been demonstrated that the Mamba-based pedestrian attribute recognition method has a reduced number of parameters and facilitates the process of fine-tuning.

\subsection{Comparison of Hybrid Mamba-Transformer} 
In this paper, we also exploit different versions of hybrid Mamba-Transformer networks for pedestrian attribute recognition, as discussed in sub-section \ref{sec:HybridMamFormer}. We will report and discuss the experimental results of these variations in this sub-section.

\begin{table}
\center
\caption{Experimental results of the proposed hybrid Mamba-Transformer networks for PAR on PETA dataset. The best and second results are highlighted in \textcolor{DarkRed}{\textbf{Red}} and \textcolor{SeaGreen4}{\textbf{Green}}, respectively. "FD": Feature Distillation, "LD": Logits Distillation} 
\label{Hybrid}  
\resizebox{0.45\textwidth}{!}{
\begin{tabular}{l|lllll}
\hline \toprule [0.5 pt]
\multicolumn{1}{l|}{\multirow{1}{*}{\textbf{Methods}}} 
& \multicolumn{1}{l}{\textbf{mA}} 
&  \multicolumn{1}{l}{\textbf{Acc}} 
&  \multicolumn{1}{l}{\textbf{Prec}} 
&  \multicolumn{1}{l}{\textbf{Recall}} 
&  \multicolumn{1}{l}{\textbf{F1}}  \\
\hline 
\textbf{ViT-B} & \textcolor{SeaGreen4}{\textbf{83.10}} & \textcolor{SeaGreen4}{\textbf{80.61}} & \textcolor{SeaGreen4}{\textbf{87.75}} & \textcolor{SeaGreen4}{\textbf{89.12}}  & \textcolor{SeaGreen4}{\textbf{88.06}} \\
\textbf{ViT-S} & 78.86 & 77.05 & 85.83 & 86.53  & 85.67 \\
\textbf{Vim-S(Baseline)} & 79.28 & 76.77 & 85.24 & 86.58  & 85.44 \\
\hline 
\textbf{(a) PaFusion}    & 75.78 & 71.74 & 81.33 & 83.57  & 81.91 \\
\textbf{(b) N-ASF}       & \textbf{81.91} & \textbf{79.54} & \textbf{86.96} & \textbf{88.44}  & \textbf{87.30} \\
\textbf{(c) ASF}         & 70.42 & 67.28 & 78.82 & 79.52  & 78.56 \\
\textbf{(d) MaFormer}    & \textbf{81.93} & \textbf{78.96} & \textbf{86.90} & \textbf{87.85}  & \textbf{86.95} \\
\textbf{(e) MaHDFT}     & \textcolor{DarkRed}{\textbf{83.64}} & \textcolor{DarkRed}{\textbf{81.11}} & \textcolor{DarkRed}{\textbf{87.90}} & \textcolor{DarkRed}{\textbf{89.66}}  & \textcolor{DarkRed}{\textbf{88.40}} \\
\textbf{(f) AdaMTF}      & 76.64 & 73.68 & 83.70 & 84.00  & 83.30  \\
\textbf{(g) KDTM (FD)}        & 77.38 & 73.96 & 83.56 & 84.13  & 83.39 \\
\textbf{(g) KDTM (LD)}        & \textbf{81.16} & \textbf{79.41} & \textbf{87.36} & \textbf{87.83}  & \textbf{87.20} \\
\textbf{(h) MaKDF (FD)}       & \textbf{80.72} & \textbf{78.57} & \textbf{86.55} & \textbf{87.61}  & \textbf{86.66} \\ 
\hline \toprule [0.5 pt] 
\end{tabular} } 
\end{table}

\begin{figure*}
    \centering
    \includegraphics[width=0.9\linewidth]{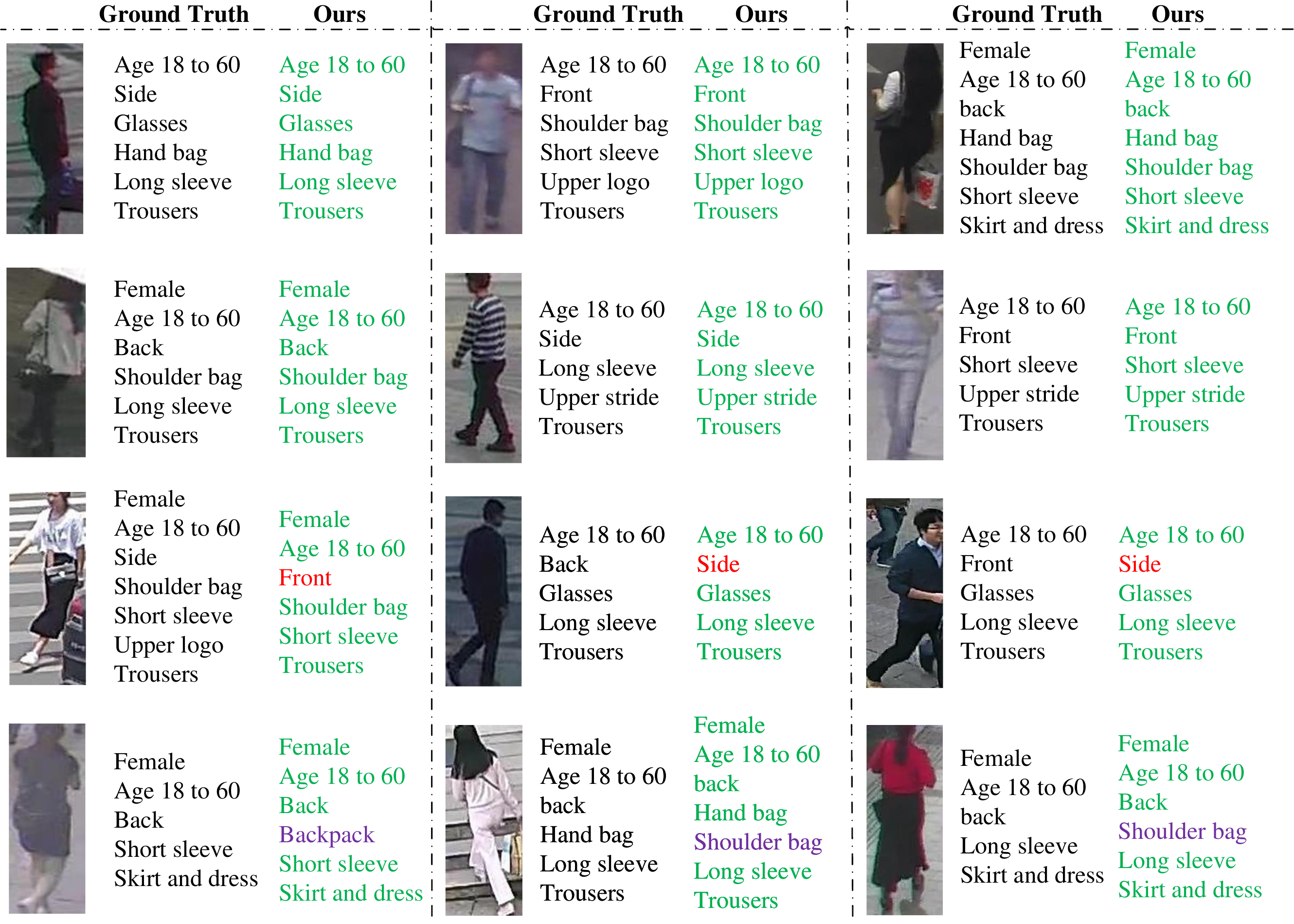}
    \caption{Visualization of the predicted human attributes using our Mamba-based PAR framework.} 
    \label{fig:PARresults_VIS}
\end{figure*}

As presented in Table \ref{Hybrid}, our analysis of the six hybrid approaches reveals that method (e) stands out as the most effective, even surpassing the performance of the most extensive Vision Transformer (ViT) models. Additionally, methods (b) and (d), while not exceeding the capabilities of ViT-B, demonstrate superior results compared to both ViT-S and Vim-S models. Regrettably, the remaining methods failed to enhance performance, which we attribute to potential design flaws or inherent limitations of the hybrid framework itself.

From the experimental results reported in Table \ref{Hybrid}, one can also find that in the method (g) KDTM, we experimented with feature distillation and Logits distillation. From the results, it seems that feature distillation has a negative effect, which may be caused by the differences between the different frameworks. Therefore, directly distilling from the predicted label distribution level can avoid the difficulty of matching the features of each layer caused by the differences between models. It can be deemed that logits-level distillation will give Vim a positive performance boost.

Drawing insights from the KDTM experiments, we refined our feature distillation technique and introduced method (h) MaKDF. This novel approach incorporates a lightweight transition module(a single-layer Vim) situated between the ViT distillation process and the Vim model itself. The output from this intermediary module is then integrated as supplementary data into the Vim backbone. As illustrated in Table \ref{Hybrid}, this refined distillation process, facilitated by a Mamba module that operates independently of the primary model structure, achieves a measure of improvement in performance over the Baseline model. Nonetheless, it has yet to surpass the efficacy of KDTM, indicating that further refinements are necessary to maximize the potential of our hybrid models.

\subsection{Visualization}   
As shown in the Fig.~\ref{fig:PARresults_VIS}, most of the prediction results of the proposed MambaPAR model based on Vim-S are better, especially in the attribute recognition of knapsack. Even if the real label is not marked, it exists from the picture, and our model can predict it. The purple label in the figure obtains the attribute. This shows that our model has better generalization performance. The main errors appear in some complex scenes and the detection of pedestrian view attributes. There is great subjective uncertainty in view attributes, which may be targeted to introduce uncertainty algorithm to solve this problem.

\subsection{Limitation Analysis}  
Although we have explored some hybrid architectures purely visually and obtained some improvements, we have not explored a suitable scheme for how to use Mamba for multi-modal fusion. From the experimental results, directly applying Mamba to modal fusion cannot obtain a consistent improvement. Therefore, how to apply Mamba to the field of multi-modal fusion still needs to be explored, and whether the hybrid architecture can also reflect the effectiveness of modal fusion is a problem that needs to be explored. We leave these issues as our future works.

\section{Conclusion} 
In this paper, our study has explored the potential of the Mamba architecture in the context of pedestrian attribute recognition (PAR) tasks. While Mamba, with its linear complexity, has shown promise in balancing accuracy and computational cost across various visual tasks, our findings suggest that its advantage in terms of memory usage is not pronounced when dealing with limited resolutions of input images in PAR tasks. The adaptation of Mamba into two typical PAR frameworks yielded mixed results; the text-image fusion approach with Vim showed enhancement potential, whereas the VMamba framework did not benefit from the interaction with attribute tags. Furthermore, our hybrid Mamba-Transformer variants demonstrated that the combination of Mamba with Transformer networks does not universally improve performance but can lead to better outcomes under specific conditions. This work serves as an empirical investigation that we hope will inspire further research into optimizing Mamba for PAR and its potential extension into multi-label recognition domains. The design and comprehensive experimentation with various network structures presented in this paper contribute valuable insights into the development of more efficient and effective models for pedestrian attribute recognition.


\small{ 
\bibliographystyle{IEEEtran}
\bibliography{reference}

\begin{thebibliography}{10}
\providecommand{\url}[1]{#1}
\csname url@samestyle\endcsname
\providecommand{\newblock}{\relax}
\providecommand{\bibinfo}[2]{#2}
\providecommand{\BIBentrySTDinterwordspacing}{\spaceskip=0pt\relax}
\providecommand{\BIBentryALTinterwordstretchfactor}{4}
\providecommand{\BIBentryALTinterwordspacing}{\spaceskip=\fontdimen2\font plus
\BIBentryALTinterwordstretchfactor\fontdimen3\font minus
  \fontdimen4\font\relax}
\providecommand{\BIBforeignlanguage}[2]{{%
\expandafter\ifx\csname l@#1\endcsname\relax
\typeout{** WARNING: IEEEtran.bst: No hyphenation pattern has been}%
\typeout{** loaded for the language `#1'. Using the pattern for}%
\typeout{** the default language instead.}%
\else
\language=\csname l@#1\endcsname
\fi
#2}}
\providecommand{\BIBdecl}{\relax}
\BIBdecl

\bibitem{wang2022PARsurvey}
X.~Wang, S.~Zheng, R.~Yang, A.~Zheng, Z.~Chen, J.~Tang, and B.~Luo,
  ``Pedestrian attribute recognition: A survey,'' \emph{Pattern Recognition},
  vol. 121, p. 108220, 2022.

\bibitem{9184102}
J.~Zhang, L.~Lin, J.~Zhu, Y.~Li, Y.-c. Chen, Y.~Hu, and S.~C.~H. Hoi,
  ``Attribute-aware pedestrian detection in a crowd,'' \emph{IEEE Transactions
  on Multimedia}, vol.~23, pp. 3085--3097, 2021.

\bibitem{zhai2024multi}
Y.~Zhai, Y.~Zeng, Z.~Huang, Z.~Qin, X.~Jin, and D.~Cao, ``Multi-prompts
  learning with cross-modal alignment for attribute-based person
  re-identification,'' \emph{The AAAI Conference on Artificial Intelligence},
  vol.~38, no.~7, pp. 6979--6987, 2024.

\bibitem{zhu2023attribute}
J.~Zhu, L.~Liu, Y.~Zhan, X.~Zhu, H.~Zeng, and D.~Tao, ``Attribute-image person
  re-identification via modal-consistent metric learning,'' \emph{International
  Journal of Computer Vision}, vol. 131, no.~11, pp. 2959--2976, 2023.

\bibitem{2015mtcnn}
A.~H. Abdulnabi, G.~Wang, J.~Lu, and K.~Jia, ``Multi-task cnn model for
  attribute prediction,'' \emph{IEEE Transactions on Multimedia}, vol.~17,
  no.~11, pp. 1949--1959, 2015.

\bibitem{wang2017JRL}
J.~Wang, X.~Zhu, S.~Gong, and W.~Li, ``Attribute recognition by joint recurrent
  learning of context and correlation,'' \emph{IEEE/CVF International
  Conference on Computer Vision}, pp. 531--540, 2017.

\bibitem{zhao2018GRL}
X.~Zhao, L.~Sang, G.~Ding, Y.~Guo, and X.~Jin, ``Grouping attribute recognition
  for pedestrian with joint recurrent learning.'' \emph{International Joint
  Conference on Artificial Intelligence}, vol. 2018, p. 27th, 2018.

\bibitem{fan2023parformer}
X.~Fan, Y.~Zhang, Y.~Lu, and H.~Wang, ``Parformer: Transformer-based multi-task
  network for pedestrian attribute recognition,'' \emph{IEEE Transactions on
  Circuits and Systems for Video Technology}, vol.~34, no.~1, pp. 411--423,
  2023.

\bibitem{cheng2022VTB}
X.~Cheng, M.~Jia, Q.~Wang, and J.~Zhang, ``A simple visual-textual baseline for
  pedestrian attribute recognition,'' \emph{IEEE Transactions on Circuits and
  Systems for Video Technology}, vol.~32, no.~10, pp. 6994--7004, 2022.

\bibitem{huang2024mambaFETrack}
J.~Huang, S.~Wang, S.~Wang, Z.~Wu, X.~Wang, and B.~Jiang, ``Mamba-fetrack:
  Frame-event tracking via state space model,'' \emph{arXiv preprint
  arXiv:2404.18174}, 2024.

\bibitem{zhu2024vim}
L.~Zhu, B.~Liao, Q.~Zhang, X.~Wang, W.~Liu, and X.~Wang, ``Vision mamba:
  Efficient visual representation learning with bidirectional state space
  model,'' \emph{arXiv preprint arXiv:2401.09417}, 2024.

\bibitem{li2024videomamba}
K.~Li, X.~Li, Y.~Wang, Y.~He, Y.~Wang, L.~Wang, and Y.~Qiao, ``Videomamba:
  State space model for efficient video understanding,'' \emph{arXiv preprint
  arXiv:2403.06977}, 2024.

\bibitem{wang2024mambaTimeSeries}
Z.~Wang, F.~Kong, S.~Feng, M.~Wang, H.~Zhao, D.~Wang, and Y.~Zhang, ``Is mamba
  effective for time series forecasting?'' \emph{arXiv preprint
  arXiv:2403.11144}, 2024.

\bibitem{wang2024SSMsurvey}
X.~Wang, S.~Wang, Y.~Ding, Y.~Li, W.~Wu, Y.~Rong, W.~Kong, J.~Huang, S.~Li,
  H.~Yang \emph{et~al.}, ``State space model for new-generation network
  alternative to transformers: A survey,'' \emph{arXiv preprint
  arXiv:2404.09516}, 2024.

\bibitem{2017pa100k}
X.~Liu, H.~Zhao, M.~Tian, L.~Sheng, J.~Shao, S.~Yi, J.~Yan, and X.~Wang,
  ``Hydraplus-net: Attentive deep features for pedestrian analysis,''
  \emph{IEEE/CVF International Conference on Computer Vision}, pp. 350--359,
  2017.

\bibitem{deng2014peta}
Y.~Deng, P.~Luo, C.~C. Loy, and X.~Tang, ``Pedestrian attribute recognition at
  far distance,'' \emph{The 22nd ACM International Conference on Multimedia},
  pp. 789--792, 2014.

\bibitem{liu2024vmamba}
Y.~Liu, Y.~Tian, Y.~Zhao, H.~Yu, L.~Xie, Y.~Wang, Q.~Ye, and Y.~Liu, ``Vmamba:
  Visual state space model,'' \emph{arXiv preprint arXiv:2401.10166}, 2024.

\bibitem{2016rapv1}
D.~{Li}, Z.~{Zhang}, X.~{Chen}, H.~{Ling}, and K.~{Huang}, ``{A Richly
  Annotated Dataset for Pedestrian Attribute Recognition},'' \emph{arXiv
  preprint arXiv:1603.07054}, 2016.

\bibitem{2019rapv2}
D.~{Li}, Z.~{Zhang}, X.~{Chen}, and K.~{Huang}, ``A richly annotated pedestrian
  dataset for person retrieval in real surveillance scenarios,'' \emph{IEEE
  Transactions on Image Processing}, vol.~28, no.~4, pp. 1575--1590, 2019.

\bibitem{li2016human}
Y.~Li, C.~Huang, C.~C. Loy, and X.~Tang, ``Human attribute recognition by deep
  hierarchical contexts,'' \emph{European Conference on Computer Vision}, pp.
  684--700, 2016.

\bibitem{2021Rethinking}
J.~{Jia}, H.~{Huang}, X.~{Chen}, and K.~{Huang}, ``{Rethinking of Pedestrian
  Attribute Recognition: A Reliable Evaluation under Zero-Shot Pedestrian
  Identity Setting},'' \emph{arXiv preprint arXiv:2107.03576}, 2021.

\bibitem{wang2023MMPTMSurvey}
X.~Wang, G.~Chen, G.~Qian, P.~Gao, X.-Y. Wei, Y.~Wang, Y.~Tian, and W.~Gao,
  ``Large-scale multi-modal pre-trained models: A comprehensive survey,''
  \emph{Machine Intelligence Research}, vol.~20, no.~4, pp. 447--482, 2023.

\bibitem{NIPS2017_3f5ee243}
A.~Vaswani, N.~Shazeer, N.~Parmar, J.~Uszkoreit, L.~Jones, A.~N. Gomez, L.~u.
  Kaiser, and I.~Polosukhin, ``Attention is all you need,'' \emph{Advances in
  Neural Information Processing Systems}, vol.~30, 2017.

\bibitem{gu2023mamba}
A.~Gu and T.~Dao, ``Mamba: Linear-time sequence modeling with selective state
  spaces,'' \emph{arXiv preprint arXiv:2312.00752}, 2023.

\bibitem{gu2021combining}
A.~Gu, I.~Johnson, K.~Goel, K.~Saab, T.~Dao, A.~Rudra, and C.~R{\'e},
  ``Combining recurrent, convolutional, and continuous-time models with linear
  state space layers,'' \emph{Advances in neural information processing
  systems}, vol.~34, pp. 572--585, 2021.

\bibitem{gu2021efficiently}
A.~Gu, K.~Goel, and C.~R{\'e}, ``Efficiently modeling long sequences with
  structured state spaces,'' \emph{arXiv preprint arXiv:2111.00396}, 2021.

\bibitem{gu2022parameterization}
A.~Gu, K.~Goel, A.~Gupta, and C.~R{\'e}, ``On the parameterization and
  initialization of diagonal state space models,'' \emph{Advances in Neural
  Information Processing Systems}, vol.~35, pp. 35\,971--35\,983, 2022.

\bibitem{gu2020hippo}
A.~Gu, T.~Dao, S.~Ermon, A.~Rudra, and C.~R{\'e}, ``Hippo: Recurrent memory
  with optimal polynomial projections,'' \emph{Advances in neural information
  processing systems}, vol.~33, pp. 1474--1487, 2020.

\bibitem{nguyen2022s4nd}
E.~Nguyen, K.~Goel, A.~Gu, G.~Downs, P.~Shah, T.~Dao, S.~Baccus, and C.~R{\'e},
  ``S4nd: Modeling images and videos as multidimensional signals with state
  spaces,'' \emph{Advances in neural information processing systems}, vol.~35,
  pp. 2846--2861, 2022.

\bibitem{dao2024transformers}
T.~Dao and A.~Gu, ``Transformers are ssms: Generalized models and efficient
  algorithms through structured state space duality,'' \emph{arXiv preprint
  arXiv:2405.21060}, 2024.

\bibitem{kalman1960new}
R.~E. Kalman, ``A new approach to linear filtering and prediction problems,''
  \emph{Journal of Basic Engineering}, vol.~82, pp. 35--45, 1960.

\bibitem{deepmar}
D.~Li, X.~Chen, and K.~Huang, ``Multi-attribute learning for pedestrian
  attribute recognition in surveillance scenarios,'' \emph{The 3rd IAPR Asian
  Conference on Pattern Recognition (ACPR)}, pp. 111--115, 2015.

\bibitem{2022drformer}
Z.~Tang and J.~Huang, ``Drformer: Learning dual relations using transformer for
  pedestrian attribute recognition,'' \emph{Neurocomputing}, vol. 497, pp.
  159--169, 2022.

\bibitem{wang2023pedestrianattributerecognitionclip}
X.~Wang, J.~Jin, C.~Li, J.~Tang, C.~Zhang, and W.~Wang, ``Pedestrian attribute
  recognition via clip based prompt vision-language fusion,'' \emph{IEEE
  Transactions on Circuits and Systems for Video Technology}, pp. 1--1, 2024.

\bibitem{kenton2019bert}
J.~D. M.-W.~C. Kenton and L.~K. Toutanova, ``Bert: Pre-training of deep
  bidirectional transformers for language understanding,'' \emph{Proceedings of
  NAACL-HLT}, pp. 4171--4186, 2019.

\bibitem{2018msvaa}
N.~Sarafianos, X.~Xu, and I.~A. Kakadiaris, ``Deep imbalanced attribute
  classification using visual attention aggregation,'' \emph{The European
  conference on computer vision}, pp. 680--697, 2018.

\bibitem{zhao2019recurrent}
X.~Zhao, L.~Sang, G.~Ding, J.~Han, N.~Di, and C.~Yan, ``Recurrent attention
  model for pedestrian attribute recognition,'' \emph{The AAAI Conference on
  Artificial Intelligence}, vol.~33, no.~01, pp. 9275--9282, 2019.

\bibitem{2019vrkd}
Q.~Li, X.~Zhao, R.~He, and K.~Huang, ``Pedestrian attribute recognition by
  joint visual-semantic reasoning and knowledge distillation,''
  \emph{International Joint Conference on Artificial Intelligence}, pp.
  833--839, 2019.

\bibitem{2019aap}
K.~Han, Y.~Wang, H.~Shu, C.~Liu, C.~Xu, and C.~Xu, ``Attribute aware pooling
  for pedestrian attribute recognition,'' \emph{International Joint Conference
  on Artificial Intelligence}, pp. 2456--2462, 2019.

\bibitem{2019VAC}
H.~Guo, K.~Zheng, X.~Fan, H.~Yu, and S.~Wang, ``Visual attention consistency
  under image transforms for multi-label image classification,'' \emph{IEEE/CVF
  Conference on Computer Vision and Pattern Recognition}, pp. 729--739, 2019.

\bibitem{2019alm}
C.~Tang, L.~Sheng, Z.-X. Zhang, and X.~Hu, ``Improving pedestrian attribute
  recognition with weakly-supervised multi-scale attribute-specific
  localization,'' \emph{IEEE/CVF International Conference on Computer Vision},
  pp. 4996--5005, 2019.

\bibitem{2020JLAC}
Z.~Tan, Y.~Yang, J.~Wan, G.~Guo, and S.~Z. Li, ``Relation-aware pedestrian
  attribute recognition with graph convolutional networks,'' \emph{The AAAI
  Conference on Artificial Intelligence}, vol.~34, no.~7, pp. 12\,055--12\,062,
  2020.

\bibitem{wu2020person}
J.~Wu, H.~Liu, J.~Jiang, M.~Qi, B.~Ren, X.~Li, and Y.~Wang, ``Person attribute
  recognition by sequence contextual relation learning,'' \emph{IEEE
  Transactions on Circuits and Systems for Video Technology}, vol.~30, no.~10,
  pp. 3398--3412, 2020.

\bibitem{2021ssc}
J.~{Jia}, X.~{Chen}, and K.~{Huang}, ``{Spatial and Semantic Consistency
  Regularizations for Pedestrian Attribute Recognition},'' \emph{arXiv preprint
  arXiv:2109.05686}, 2021.

\bibitem{2022iaacaps}
J.~Wu, Y.~Huang, Z.~Gao, Y.~Hong, J.~Zhao, and X.~Du, ``Inter-attribute
  awareness for pedestrian attribute recognition,'' \emph{Pattern Recognition},
  vol. 131, p. 108865, 2022.

\bibitem{Chen2022MCFL}
L.~C. andJingkuan Song andXuerui Zhang~andMingsheng Shang, ``Mcfl: multi-label
  contrastive focal loss for deep imbalanced pedestrian attribute
  recognition,'' \emph{Neural Computing and Applications}, vol.~34, no.~19, pp.
  16\,701--16\,715, 2022.

\bibitem{guo2022visual}
H.~Guo, X.~Fan, and S.~Wang, ``Visual attention consistency for human attribute
  recognition,'' \emph{International Journal of Computer Vision}, vol. 130,
  no.~4, pp. 1088--1106, 2022.

\bibitem{jia2022learning}
J.~Jia, N.~Gao, F.~He, X.~Chen, and K.~Huang, ``Learning disentangled attribute
  representations for robust pedestrian attribute recognition,'' \emph{The AAAI
  Conference on Artificial Intelligence}, vol.~36, no.~1, pp. 1069--1077, 2022.

\bibitem{Fan2022CGCN}
H.~Fan, H.-M. Hu, S.~Liu, W.~Lu, and S.~Pu, ``Correlation graph convolutional
  network for pedestrian attribute recognition,'' \emph{IEEE Transactions on
  Multimedia}, vol.~24, pp. 49--60, 2022.

\bibitem{yang2021cascaded}
Y.~Yang, Z.~Tan, P.~Tiwari, H.~M. Pandey, J.~Wan, Z.~Lei, G.~Guo, and S.~Z. Li,
  ``Cascaded split-and-aggregate learning with feature recombination for
  pedestrian attribute recognition,'' \emph{International Journal of Computer
  Vision}, vol. 129, no.~10, pp. 2731--2744, 2021.

\bibitem{lu2023oagcn}
W.-Q. Lu, H.-M. Hu, J.~Yu, Y.~Zhou, H.~Wang, and B.~Li, ``Orientation-aware
  pedestrian attribute recognition based on graph convolution network,''
  \emph{IEEE Transactions on Multimedia}, vol.~26, pp. 28--40, 2024.

\bibitem{jin2023sequencepar}
J.~Jin, X.~Wang, C.~Li, L.~Huang, and J.~Tang, ``Sequencepar: Understanding
  pedestrian attributes via a sequence generation paradigm,'' \emph{arXiv
  preprint arXiv:2312.01640}, 2023.

\bibitem{SHEN2024110194}
J.~Shen, T.~Guo, X.~Zuo, H.~Fan, and W.~Yang, ``Sspnet: Scale and spatial
  priors guided generalizable and interpretable pedestrian attribute
  recognition,'' \emph{Pattern Recognition}, vol. 148, p. 110194, 2024.

\bibitem{jin2024pedestrianattributerecognitionnew}
J.~Jin, X.~Wang, Q.~Zhu, H.~Wang, and C.~Li, ``Pedestrian attribute
  recognition: A new benchmark dataset and a large language model augmented
  framework,'' \emph{arXiv preprint arXiv:2408.09720}, 2024.

\bibitem{ZHENG2023105708}
A.~Zheng, H.~Wang, J.~Wang, H.~Huang, R.~He, and A.~Hussain, ``Diverse features
  discovery transformer for pedestrian attribute recognition,''
  \emph{Engineering Applications of Artificial Intelligence}, vol. 119, p.
  105708, 2023.

\bibitem{Zhou2023Co}
Y.~Zhou, H.-M. Hu, J.~Yu, Z.~Xu, W.~Lu, and Y.~Cao, ``A solution to
  co-occurence bias: attributes disentanglement via mutual information
  minimization for pedestrian attribute recognition,'' \emph{International
  Joint Conference on Artificial Intelligence}, pp. 1831--1839, 2023.

\bibitem{zuo2023plip}
J.~Zuo, C.~Yu, N.~Sang, and C.~Gao, ``Plip: Language-image pre-training for
  person representation learning,'' \emph{arXiv preprint arXiv:2305.08386},
  2023.

\bibitem{adam2015}
S.~J. Reddi, S.~Kale, and S.~Kumar, ``On the convergence of adam and beyond,''
  \emph{arXiv preprint arXiv:1904.09237}, 2019.

\bibitem{paszke2019pytorch}
A.~Paszke, S.~Gross, F.~Massa, A.~Lerer, J.~Bradbury, G.~Chanan, T.~Killeen,
  Z.~Lin, N.~Gimelshein, L.~Antiga \emph{et~al.}, ``Pytorch: An imperative
  style, high-performance deep learning library,'' \emph{Advances in neural
  information processing systems}, vol.~32, 2019.

\bibitem{radford2021CLIP}
A.~Radford, J.~W. Kim, C.~Hallacy, A.~Ramesh, G.~Goh, S.~Agarwal, G.~Sastry,
  A.~Askell, P.~Mishkin, J.~Clark \emph{et~al.}, ``Learning transferable visual
  models from natural language supervision,'' \emph{International Conference on
  Machine Learning}, pp. 8748--8763, 2021.

\bibitem{2015rcnn}
G.~Gkioxari, R.~Girshick, and J.~Malik, ``Contextual action recognition with
  r*cnn,'' \emph{IEEE/CVF International Conference on Computer Vision}, pp.
  1080--1088, 2015.

\bibitem{2017srn}
F.~{Zhu}, H.~{Li}, W.~{Ouyang}, N.~{Yu}, and X.~{Wang}, ``{Learning Spatial
  Regularization with Image-level Supervisions for Multi-label Image
  Classification},'' \emph{arXiv preprint arXiv:1702.05891}, 2017.

\bibitem{2019dahar}
M.~{Wu}, D.~{Huang}, Y.~{Guo}, and Y.~{Wang}, ``{Distraction-Aware Feature
  Learning for Human Attribute Recognition via Coarse-to-Fine Attention
  Mechanism},'' \emph{arXiv preprint arXiv:1911.11351}, 2019.

\end{thebibliography}
}

\end{document}